\documentclass[11pt]{article}
\usepackage{graphicx} 

\usepackage{natbib}
\usepackage{multirow}
\usepackage{caption}
\usepackage{siunitx}
\usepackage{ifthen}
\newboolean{isSingleColumn}

\makeatletter
\if@twocolumn
    \setboolean{isSingleColumn}{false}
\else
    \setboolean{isSingleColumn}{true}
\fi

\newcommand{\adaptiveHeader}[1]{%
    \ifthenelse{\boolean{isSingleColumn}}{%
        \paragraph{#1} 
    }{%
        \textbf{#1}    
    }%
}
\usepackage{booktabs} 
\usepackage[ruled]{algorithm2e} 

\usepackage{fullpage}
\usepackage{authblk}
\usepackage{tikz}
\usepackage{mathtools}
\usepackage{amsmath, amssymb, amsfonts}
\usepackage{hyperref}
\usepackage{tikz}
\usepackage{bbding}
\usepackage[capitalize,noabbrev]{cleveref}
\usepackage{subcaption} 
\usetikzlibrary{
    decorations.pathmorphing, 
    arrows.meta,             
    positioning,             
    calc
}
\usetikzlibrary{shapes, arrows.meta, positioning, fit, backgrounds, decorations.pathreplacing, calc, shadows}

\definecolor{patientblue}{RGB}{70, 130, 180}
\definecolor{donorred}{RGB}{220, 80, 80}
\definecolor{clustergray}{RGB}{240, 240, 245}
\definecolor{edgegreen}{RGB}{46, 139, 87}

\tikzset{
    complexheart/.pic={
        \begin{scope}[yscale=-1, shift={(-4.5, -9)}]
            \draw[fill=red!30!white](.456,3.236)
                .. controls (.422,4.168) and (.408,5.095) .. (.461,6.046)
                .. controls (.475,6.228) and (.365,6.400) .. (.379,6.601)
                -- (.819,11.816)
                .. controls (.843,12.194) and (.838,13.389) .. (.900,13.972)
                .. controls (.943,14.340) and (2.870,14.340) .. (2.903,13.972)
                .. controls (2.927,13.699) and (2.932,13.436) .. (2.903,13.169)
                -- (1.847,5.401)
                .. controls (1.914,4.627) and (2.033,3.924) .. (2.129,3.193)
                .. controls (2.177,2.906) and (.441,2.863) .. (.456,3.236) ;
            \draw[fill=red!30!white] (.456,3.236)
                .. controls (.422,4.168) and (.408,5.095) .. (.461,6.046)
                .. controls (.475,6.228) and (.365,6.400) .. (.379,6.601)
                -- (.819,11.816)
                .. controls (.843,12.194) and (.838,13.389) .. (.900,13.972)
                .. controls (.943,14.340) and (2.870,14.340) .. (2.903,13.972)
                .. controls (2.927,13.699) and (2.932,13.436) .. (2.903,13.169)
                -- (1.847,5.401)
                .. controls (1.914,4.627) and (2.033,3.924) .. (2.129,3.193)
                .. controls (2.177,2.906) and (.441,2.863) .. (.456,3.236) ;
            \draw[fill=red!60!white] (.303,6.697)
                .. controls (-.161,7.442) and (-.022,9.641) .. (.484,11.118)
                .. controls (.795,12.022) and (1.297,12.414) .. (2.062,12.801)
                .. controls (3.429,13.513) and (5.112,13.714) .. (6.598,13.795)
                .. controls (9.495,13.957) and (9.738,11.735) .. (8.974,8.967)
                .. controls (8.744,8.126) and (8.123,7.815) .. (7.917,7.222)
                .. controls (7.683,6.606) and (7.310,5.999) .. (6.598,5.420)
                .. controls (5.570,4.579) and (1.230,5.210) .. (.303,6.697) ;
            \fill[fill=red!25!white] (2.540,6.405)
                .. controls (2.296,6.257) and (1.890,6.300) .. (1.608,6.424)
                .. controls (1.274,6.577) and (1.101,6.802) .. (1.001,7.151)
                .. controls (.858,7.652) and (1.025,8.293) .. (1.106,8.795)
                .. controls (1.149,9.072) and (1.149,9.402) .. (1.254,9.684)
                .. controls (1.278,9.746) and (1.278,9.823) .. (1.360,9.870)
                .. controls (1.426,9.909) and (1.412,9.732) .. (1.460,9.651)
                .. controls (1.747,9.144) and (2.210,8.346) .. (2.502,7.834)
                .. controls (2.602,7.657) and (2.631,7.437) .. (2.531,7.303)
                .. controls (2.411,7.146) and (2.206,6.883) .. (2.306,6.768)
                .. controls (2.382,6.682) and (2.621,6.453) .. (2.540,6.405) ;
            \fill[fill=red!25!white] (4.204,6.606)
                .. controls (5.413,6.434) and (5.327,7.543) .. (5.466,7.796)
                .. controls (5.599,8.054) and (6.350,8.785) .. (6.498,9.029)
                .. controls (6.646,9.273) and (7.282,9.947) .. (7.067,10.616)
                .. controls (6.847,11.290) and (5.776,11.582) .. (5.212,11.429)
                .. controls (4.643,11.271) and (3.773,10.941) .. (3.166,11.113)
                .. controls (2.559,11.280) and (1.350,10.793) .. (1.790,10.052)
                .. controls (2.225,9.311) and (3.640,6.692) .. (4.204,6.606) ;
            \draw[fill=yellow] (1.307,3.107)
                .. controls (1.675,3.107) and (1.976,3.169) .. (1.976,3.250)
                .. controls (1.976,3.327) and (1.675,3.394) .. (1.307,3.394)
                .. controls (.934,3.394) and (.633,3.327) .. (.633,3.250)
                .. controls (.633,3.169) and (.934,3.107) .. (1.307,3.107) ;
            \draw[fill=blue!20!red!80!white] (1.307,3.107)
                .. controls (1.675,3.107) and (1.976,3.169) .. (1.976,3.250)
                .. controls (1.976,3.327) and (1.675,3.394) .. (1.307,3.394)
                .. controls (.934,3.394) and (.633,3.327) .. (.633,3.250)
                .. controls (.633,3.169) and (.934,3.107) .. (1.307,3.107) ;
            \fill[fill=red!10!white] (1.024,14.000)
                .. controls (.991,13.929) and (.924,12.146) .. (1.039,12.323)
                .. controls (1.120,12.452) and (1.498,12.753) .. (1.870,12.767)
                .. controls (2.004,12.777) and (1.918,14.129) .. (1.875,14.163)
                .. controls (1.789,14.230) and (1.110,14.173) .. (1.024,14.000) ;
            \fill[fill=red!10!white] (.618,3.432)
                -- (.618,6.247)
                .. controls (.852,5.941) and (1.086,5.745) .. (1.345,5.578)
                .. controls (1.364,5.033) and (1.368,4.125) .. (1.345,3.556)
                .. controls (1.096,3.556) and (.862,3.518) .. (.618,3.432) ;
            \draw[fill=white] (4.872,5.215)
                .. controls (4.944,5.607) and (4.891,5.884) .. (4.595,6.061)
                .. controls (4.389,6.185) and (3.749,6.013) .. (3.864,5.664)
                .. controls (3.974,5.339) and (4.088,5.081) .. (4.346,4.923)
                .. controls (4.533,4.813) and (4.858,5.119) .. (4.872,5.215) ;
            
            \draw[fill=red!80!blue] (5.662,5.665)
                .. controls (5.643,5.732) and (5.624,5.803) .. (5.604,5.880)
                .. controls (5.944,6.076) and (6.355,6.640) .. (6.522,7.314)
                .. controls (6.699,8.031) and (6.336,8.284) .. (5.853,8.796)
                .. controls (5.523,9.154) and (5.313,9.541) .. (5.303,10.014)
                .. controls (5.485,9.106) and (5.844,8.958) .. (6.030,8.877)
                .. controls (6.278,8.767) and (6.221,9.221) .. (6.159,10.029)
                .. controls (6.341,9.412) and (6.298,8.944) .. (6.274,8.752)
                .. controls (6.245,8.561) and (6.589,8.370) .. (6.685,8.093)
                .. controls (6.799,7.777) and (6.924,8.069) .. (6.991,8.208)
                .. controls (7.153,8.547) and (7.311,8.681) .. (7.263,8.987)
                .. controls (7.096,10.033) and (6.780,10.387) .. (6.250,10.650)
                .. controls (6.728,10.492) and (6.833,10.373) .. (6.976,10.230)
                .. controls (6.991,10.440) and (6.900,10.870) .. (6.962,11.209)
                .. controls (6.957,10.669) and (7.277,9.197) .. (7.469,9.044)
                .. controls (7.579,8.953) and (8.296,9.632) .. (8.372,10.139)
                .. controls (8.458,10.698) and (8.936,11.535) .. (8.716,12.213)
                .. controls (8.970,11.458) and (8.535,10.636) .. (8.635,10.445)
                .. controls (8.778,10.641) and (9.060,11.037) .. (9.027,11.257)
                .. controls (8.994,11.477) and (9.170,11.874) .. (9.094,12.333)
                .. controls (9.228,11.883) and (9.065,11.702) .. (9.118,11.377)
                .. controls (9.166,11.052) and (8.979,10.626) .. (8.836,10.359)
                .. controls (8.549,9.828) and (7.889,9.187) .. (7.373,8.595)
                .. controls (7.708,8.796) and (8.535,9.077) .. (8.487,9.369)
                .. controls (8.434,9.661) and (8.860,9.900) .. (9.027,10.306)
                .. controls (8.903,9.928) and (8.587,9.656) .. (8.568,9.321)
                .. controls (8.554,8.982) and (8.023,8.815) .. (7.889,8.762)
                .. controls (7.995,8.752) and (8.253,8.748) .. (8.348,8.834)
                .. controls (8.444,8.920) and (8.702,9.097) .. (8.912,9.044)
                .. controls (8.539,9.044) and (8.463,8.671) .. (8.286,8.657)
                .. controls (8.109,8.647) and (7.641,8.566) .. (7.531,8.470)
                .. controls (7.015,8.016) and (6.599,7.223) .. (6.819,7.271)
                .. controls (7.344,7.381) and (7.899,7.471) .. (8.377,7.988)
                .. controls (8.023,7.567) and (7.999,7.562) .. (7.478,7.314)
                .. controls (7.091,7.137) and (6.537,6.979) .. (6.355,6.511)
                .. controls (6.250,6.229) and (5.887,5.789) .. (5.662,5.665) ;
            \draw[fill=red!80!blue] (3.387,5.951)
                .. controls (3.779,7.003) and (2.665,8.600) .. (1.857,9.866)
                .. controls (1.384,10.607) and (1.704,11.157) .. (2.364,11.305)
                .. controls (2.842,11.410) and (4.515,11.004) .. (4.993,11.023)
                .. controls (5.480,11.037) and (5.604,10.856) .. (5.882,10.559)
                .. controls (5.815,10.894) and (5.256,11.305) .. (4.997,11.209)
                .. controls (4.735,11.114) and (4.199,11.229) .. (3.975,11.338)
                .. controls (4.280,11.425) and (4.787,11.401) .. (5.179,11.601)
                .. controls (5.571,11.802) and (5.920,11.750) .. (6.159,11.300)
                .. controls (6.140,11.730) and (5.538,11.955) .. (5.126,11.783)
                .. controls (4.715,11.616) and (3.736,11.434) .. (3.401,11.434)
                .. controls (3.176,11.429) and (3.009,11.468) .. (2.617,11.563)
                .. controls (3.324,11.635) and (4.252,12.084) .. (5.165,12.304)
                .. controls (5.863,12.471) and (6.680,12.452) .. (6.790,12.189)
                .. controls (6.895,11.946) and (7.072,11.840) .. (7.473,11.793)
                .. controls (7.000,12.008) and (6.866,12.156) .. (6.814,12.438)
                .. controls (7.015,12.490) and (7.411,12.256) .. (7.626,12.457)
                .. controls (7.846,12.658) and (8.439,12.653) .. (8.864,12.381)
                .. controls (8.386,12.730) and (7.770,12.710) .. (7.607,12.557)
                .. controls (7.445,12.405) and (7.077,12.553) .. (6.862,12.591)
                .. controls (6.647,12.634) and (6.446,12.644) .. (5.891,12.548)
                .. controls (5.714,12.620) and (5.657,12.715) .. (5.867,12.792)
                .. controls (6.082,12.835) and (6.537,12.806) .. (7.014,12.806)
                .. controls (7.258,12.806) and (7.430,12.935) .. (7.631,12.811)
                .. controls (7.832,12.682) and (8.033,12.873) .. (8.324,13.016)
                .. controls (7.999,12.906) and (7.746,12.768) .. (7.660,12.897)
                .. controls (7.574,13.031) and (7.263,12.916) .. (7.163,12.940)
                .. controls (7.292,12.992) and (7.493,12.935) .. (7.507,13.136)
                .. controls (7.526,13.337) and (8.057,13.284) .. (7.999,13.652)
                .. controls (7.923,13.294) and (7.540,13.423) .. (7.406,13.179)
                .. controls (7.277,12.940) and (6.885,13.055) .. (6.680,13.031)
                .. controls (6.250,12.973) and (5.557,13.074) .. (5.461,12.639)
                .. controls (5.217,12.600) and (4.921,12.514) .. (4.682,12.419)
                .. controls (4.687,12.538) and (5.103,13.026) .. (5.461,13.102)
                .. controls (5.815,13.174) and (6.532,13.155) .. (6.699,13.681)
                .. controls (6.345,13.174) and (5.614,13.356) .. (5.294,13.260)
                .. controls (4.978,13.160) and (4.213,12.285) .. (3.898,12.189)
                .. controls (3.583,12.099) and (3.272,11.922) .. (3.157,11.989)
                .. controls (3.062,12.084) and (3.119,12.600) .. (3.583,12.581)
                .. controls (4.046,12.562) and (4.362,12.882) .. (4.620,13.356)
                .. controls (4.271,12.897) and (3.721,12.615) .. (3.358,12.744)
                .. controls (2.990,12.868) and (2.823,11.936) .. (2.598,11.812)
                .. controls (2.373,11.683) and (1.594,11.377) .. (1.503,11.042)
                .. controls (1.465,11.377) and (2.106,11.903) .. (2.043,12.677)
                .. controls (1.953,11.922) and (1.522,11.783) .. (1.317,11.377)
                .. controls (1.111,10.966) and (1.212,10.555) .. (1.417,9.991)
                .. controls (1.422,9.971) and (3.836,6.267) .. (2.588,5.985)
                .. controls (2.235,5.904) and (1.981,5.813) .. (1.757,5.660)
                .. controls (1.871,4.489) and (2.187,2.481) .. (2.650,2.405)
                .. controls (3.176,2.314) and (4.185,2.878) .. (4.175,3.289)
                .. controls (4.166,3.724) and (3.817,4.766) .. (3.994,5.201)
                .. controls (3.759,5.397) and (3.277,5.617) .. (3.387,5.951) ;
            \fill[fill=red!90!blue!70!white] (2.942,5.378)
                .. controls (2.875,5.339) and (2.736,5.378) .. (2.722,5.459)
                .. controls (2.712,5.535) and (2.741,5.583) .. (2.779,5.674)
                .. controls (3.013,6.066) and (3.200,6.716) .. (2.980,7.266)
                .. controls (2.803,7.720) and (2.564,8.226) .. (2.296,8.661)
                .. controls (2.048,9.068) and (1.627,9.670) .. (1.565,10.363)
                .. controls (1.579,10.105) and (1.837,9.737) .. (1.952,9.512)
                .. controls (2.263,8.905) and (3.071,7.815) .. (3.214,7.180)
                .. controls (3.362,6.534) and (3.243,5.875) .. (3.071,5.597)
                .. controls (3.066,5.540) and (3.028,5.425) .. (2.942,5.378) ;
            \fill[fill=red!90!blue!70!white] (2.793,2.424)
                .. controls (2.253,3.542) and (2.115,3.948) .. (2.014,5.712)
                .. controls (2.019,5.784) and (2.349,5.903) .. (2.339,5.841)
                .. controls (2.301,4.665) and (2.650,3.289) .. (3.138,2.486)
                .. controls (3.032,2.447) and (2.870,2.419) .. (2.793,2.424) ;
            \draw (2.526,6.104)
                .. controls (2.990,5.937) and (2.541,6.688) .. (2.545,6.673)
                .. controls (2.454,6.831) and (2.521,6.869) .. (2.564,6.922)
                .. controls (2.717,7.123) and (2.932,7.165) .. (2.660,7.471) ;
            \draw (.461,3.241)
                .. controls (.576,3.657) and (2.187,3.657) .. (2.129,3.198) ;
            \draw[fill=red!30!white] (3.687,5.731)
                .. controls (3.520,6.229) and (5.541,6.616) .. (5.556,6.214)
                .. controls (5.570,5.827) and (5.790,5.325) .. (5.800,4.919)
                .. controls (5.809,4.584) and (4.480,4.250) .. (4.308,4.517)
                .. controls (4.084,4.866) and (3.854,5.249) .. (3.687,5.731) ;
            \draw (4.299,4.531)
                .. controls (4.189,4.885) and (5.752,5.315) .. (5.800,4.961) ;
            \draw[fill=blue!20!red!80!white] (5.097,4.594)
                .. controls (5.436,4.694) and (5.699,4.823) .. (5.680,4.919)
                .. controls (5.656,5.009) and (5.360,5.000) .. (5.020,4.899)
                .. controls (4.681,4.799) and (4.428,4.675) .. (4.452,4.584)
                .. controls (4.475,4.493) and (4.757,4.493) .. (5.097,4.594) ;
        \end{scope}
    }
}
\hypersetup{
    colorlinks=true,
    linkcolor=blue,
    filecolor=magenta,      
    urlcolor=cyan,
}

\usepackage{xcolor}

\newtheorem{theorem}{Theorem}
\newtheorem{proposition}[theorem]{Proposition}

\title{Learning Potentials for Dynamic Matching and Application to Heart Transplantation}

\author[1]{Itai Zilberstein\thanks{Correspondence to \texttt{sandholm@cs.cmu.edu}}}
\author[1]{Ioannis Anagnostides}
\author[2]{Zachary W. Sollie}
\author[2]{Arman Kilic}
\author[1,3]{Tuomas Sandholm}

\affil[1]{Department of Computer Science, Carnegie Mellon University, Pittsburgh, PA}
\affil[2]{Department of Surgery, Division of Cardiothoracic Surgery, Medical University of South Carolina, Charleston, SC}
\affil[3]{Additional affiliations: Strategy Robot, Inc., Strategic Machine, Inc., Optimized Markets, Inc.}

\begin{document}

\maketitle
\thispagestyle{empty}

\begin{abstract}
Each year, thousands of patients in need of heart transplants face life-threatening wait times due to organ scarcity. While allocation policies aim to maximize population-level outcomes, current approaches often fail to account for the dynamic arrival of organs and the composition of waitlisted candidates, thereby hampering efficiency. The United States is transitioning from rigid, rule-based allocation to more flexible data-driven models. In this paper, we propose a novel framework for non-myopic policy optimization in general online matching relying on \textit{potentials}, a concept originally introduced for kidney exchange. We develop scalable and accurate ways of learning potentials that are higher-dimensional and more expressive than prior approaches. Our approach is a form of \textit{self-supervised imitation learning}: the potentials are trained to mimic an omniscient algorithm that has perfect foresight. We focus on the application of heart transplant allocation and demonstrate, using real historical data, that our policies significantly outperform prior approaches---including the current US \textit{status quo} policy and the proposed \textit{continuous distribution} framework---in optimizing for population-level outcomes. Our analysis and methods come at a pivotal moment in US policy, as the current heart transplant allocation system is under review. We propose a scalable and theoretically grounded path toward more effective organ allocation.
\end{abstract}
 
\newpage
\setcounter{page}{1}

\section{Introduction}

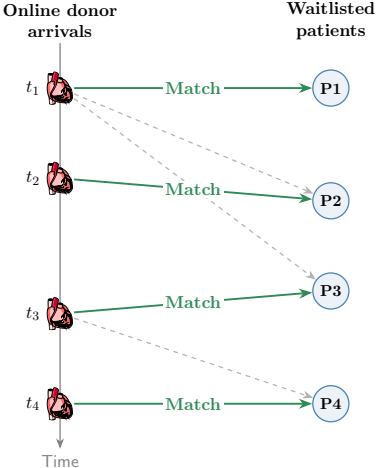
\begin{figure}[b!]
    \centering
    \scalebox{0.6}{\begin{tikzpicture}[
    font=\sffamily,
    >=Stealth,
    patient/.style={circle, draw=patientblue, fill=patientblue!10, thick, inner sep=0pt, minimum size=0.8cm, font=\bfseries\small},
    donor/.style={inner sep=0pt, minimum size=0pt},
    selected_match/.style={->, very thick, edgegreen},
    potential_match/.style={->, dashed, gray!60, thick},
    edge_label/.style={midway, fill=white, inner sep=1.5pt, text=edgegreen, font=\bfseries}
]

\draw[->, thick, gray] (-3, 4.5) -- (-3, -4.5) node[below] {Time};
\node[align=center, font=\bfseries] at (-3, 5) {Online donor\\arrivals};

\node[donor, label={left:$t_1$}] (d1) at (-3, 3.5)  {\tikz{\pic[scale=0.06]{complexheart};}};
\node[donor, label={left:$t_2$}] (d2) at (-3, 1.5)  {\tikz{\pic[scale=0.06]{complexheart};}};
\node[donor, label={left:$t_3$}] (d3) at (-3, -1.5) {\tikz{\pic[scale=0.06]{complexheart};}};
\node[donor, label={left:$t_4$}] (d4) at (-3, -3.5) {\tikz{\pic[scale=0.06]{complexheart};}};

\node[align=center, font=\bfseries] at (3, 5) {Waitlisted\\patients};

\node[patient] (p1) at (3, 3.5) {P1};
\node[patient] (p2) at (3, 1.0) {P2};
\node[patient] (p3) at (3, -1.0) {P3};
\node[patient] (p4) at (3, -3.5) {P4};


\draw[selected_match] (d1) -- (p1) node[edge_label] {Match};
\draw[potential_match] (d1) -- (p2); 
\draw[potential_match] (d1) -- (p3); 

\draw[selected_match] (d2) -- (p2) node[edge_label] {Match};

\draw[selected_match] (d3) -- (p3) node[edge_label] {Match};

\draw[selected_match] (d4) -- (p4) node[edge_label] {Match};

\draw[potential_match] (d3) -- (p4);

\end{tikzpicture}}
    \caption{Schematic illustration of online heart transplant allocation.}
    \label{fig:heart-matching}
\end{figure}

Organ transplantation is the preferred treatment for patients with end-stage heart failure. However, the demand for life-saving organs far exceeds the available supply, requiring an efficient and equitable allocation policy~\citep{Cameli22:Donor}. The difficulty of heart transplant allocation arises from its dynamic and time-sensitive nature. Heart transplantation depends exclusively on deceased donors. Since hearts have a limited cold ischemia time, they must be allocated and transplanted quickly upon availability. Decision-makers need to irrevocably match arriving organs with waiting patients without knowledge of future arrivals (\Cref{fig:heart-matching}). The system involves a challenging tradeoff: accepting a marginal match for a patient may prevent them from receiving a superior match later on. Conversely, waiting for a better organ risks the patient dying on the waitlist. Organ allocation is arguably the highest-stakes application of online matching. 

Following a major revision in 2018, the US heart transplant allocation policy is again  transitioning to new solutions. The current system is a rigid, hierarchical tier-based system. The 2018 revision expanded a 3-tier system to a 6-tier one to better quantify the urgency of candidates~\citep{Kilic21:Evolving}. However, the current system has a major limitation: it fails to leverage granular estimates of pretransplant mortality and post-transplant outcomes, often conflating heterogeneous patients within the same priority status~\citep{Shore20:Changes,Zhang24:Development}. This results in the system allocating scarce organs to patients with worse outcomes~\citep{Cogswell20:Early}. To overcome these deficiencies, the \textit{Organ Procurement and Transplantation Network (OPTN)} is reviewing the proposed \textit{continuous distribution} framework~\citep{OPTN2024:HeartUpdate, Papalexopoulos24:Reshaping}. Continuous distribution, which is the current allocation scheme for lung transplants in the US, assigns each patient a real-valued score. While continuous distribution replaces rigid tiers with a more flexible score, it assumes that the score is a static weighted sum of patient feature values. This underlying class of policies is overly restrictive. Even with optimized weights, it lacks the expressiveness to consider patient-donor feature interactions and to fully adapt to the complex, non-linear dynamics of arriving organs and an evolving waitlist. Designing an allocation policy that maximizes population-level outcomes in this complicated setting is a key challenge.


\subsection{Our contributions}

In this paper, we propose a computational approach for designing a matching policy and show on real data that it significantly outperforms prior approaches and proposals.


Our methodology builds on the notion of \textit{potentials}, which were introduced for kidney exchange~\citep{Dickerson12:Dynamic,Dickerson15:Futurematch}. Kidney exchange is a different problem than online matching as it involves cycles longer than two~\citep{Roth04:Kidney,Abraham07:Clearing} and never-ending altruist donor chains~\citep{Rees09:Nonsimultaneous,Dickerson19:Failure,Dickerson16:Position}. Potentials enable a policy to account for long-term pool composition by offsetting edge weights to reflect far-out utility rather than near-term gains. Those two prior potentials-based papers used \textit{Sequential Model-based Algorithm Configuration (SMAC)}~\citep{Hutter11:Sequential} to learn low-dimensional linear potential functions. Preliminary work has shown that simple linear potentials outperform myopic policies in heart allocation---which is a dynamic matching problem---as well~\citep{Anagnostides25:Policy}. However, as we will show, simple linear parameterizations are insufficient to capture the full complexity of heart transplant allocations. Unlike kidney exchange, which allows for accumulation of the live donors, heart transplants require immediate, irrevocable allocations of deceased donors. Each match must be made with little foresight of future donor arrivals, and a single decision may have long-term consequences on population outcomes. As we will show, black-box optimization is insufficient for learning complex, non-linear potentials required for achieving better outcomes in heart transplant allocation.

To bypass the inefficiencies of black-box optimization, we formulate a novel approach for learning potential functions using offline self-supervised imitation learning. We develop an omniscient oracle that identifies optimal allocations in hindsight using full knowledge of donor arrivals over a historical period. Our policy optimization then learns potential functions that cause our policy to imitate the omniscient oracle.

Our framework not only improves the way potentials are learned but also enables the learning of more expressive potentials. We enable learning high-dimensional and non-linear functions via self-supervised learning. These potential functions can reflect complex interactions between patient characteristics, donor features, the arrival distribution, and the waitlist state.

While our approach is applicable to general online matching problems, we focus on the application of heart transplant allocation. We demonstrate using real historical data from the \textit{United Network for Organ Sharing (UNOS)} that our approach considerably outperforms all prior approaches and proposals in optimizing for population-level outcomes, including the current US \textit{status quo} policy and the proposed continuous distribution framework. To our knowledge, our work is the first to thoroughly evaluate the continuous distribution framework from an optimization standpoint. We further demonstrate that our framework can effectively optimize weights within the continuous distribution system, offering an alternative to current committee-assigned or heuristic weight-selection methods. Finally, our best potential-based policy is near-optimal: it yields 95\% of the solution quality upper bound that we generate using omniscient matching.

We also address data scarcity to reduce overfitting inherent in historical simulations. We introduce a data augmentation pipeline that generates diverse, medically realistic training scenarios from historical trajectories. Our method randomizes the temporal arrival of donors and patients while preserving real clinical features and progressions. The augmentation helps the model learn potentials that capture the underlying matching logic rather than overfitting to historical timestamps.

In summary, we develop a new framework to harness the power of data-driven learning within online matching problems. We apply our framework to develop significantly better policies for heart transplant allocation that account for the long-term consequences of matching scarce resources. Our methods and experiments come at a pivotal moment in US policy, as the current heart transplant allocation system is under review. 

\subsection{Additional related work}

\paragraph{Organ allocation} Research in data-driven organ allocation typically restricts optimization to simplified, interpretable policy classes. Early research proposed using techniques from queueing theory to maximize welfare in the context of kidney allocation~\citep{Su04:Patient}. More recently, policies for kidney allocation were proposed based on a point system that ranks patients according to fixed priority criteria~\citep{Bertsimas13:Fairness}. Similarly, the current \textit{status quo} policy for adult heart transplants in the US utilizes a rigid six-tier hierarchy~\citep{Kilic21:Evolving}. The system is widely criticized for failing to optimize for pretransplant and post-transplant outcomes~\citep{Shore20:Changes}. To address these limitations, the OPTN is reviewing transitioning to the \textit{continuous distribution framework}~\citep{Papalexopoulos24:Reshaping}. Continuous distribution has already been deployed for the allocation of lungs, and is being considered for hearts, kidneys, livers, intestines, and pancreases~\citep{OPTN25:ContinuousDistribution}. That framework assigns a \textit{composite allocation score (CAS)} to each patient, a linear combination of features such as medical urgency and geography. While CAS allows for more granularity than tiers, it ultimately reduces the allocation problem to optimizing a small, fixed set of weights. As of early 2026, these weights remain unfinalized for hearts, yet a preliminary feature set is available~\citep{Cummiskey25:Understanding,OPTN2024:HeartUpdate}. In addition to evaluating our own potential-based policies, we apply our framework to derive optimized weights for CAS so we can compare our approach against theirs.

Parallel research focuses on improving the estimation of outcomes of possible matches rather than on optimizing the matching policy~\citep{Lee18:Deephit, Nilsson15:International,Ayers21:Using}. The \textit{OrganITE} framework~\citep{Berrevoets20:OrganITE} prioritizes matches based on estimated individual treatment effects and organ scarcity. Its successor, \textit{OrganSync}~\citep{Berrevoets21:Learning}, groups organs into clustered types and employs priority queues for each type, with the goal of obtaining more robust counterfactual survival estimates. In contrast, our work focuses on the underlying matching logic. We utilize a simulator based on real UNOS data that is agnostic to the underlying predictive model of transplant outcomes. This ensures our policy optimization remains compatible as medical prognosis tools evolve. 

Finally, it is worth noting that incentive compatibility has been studied in kidney exchange (\emph{e.g.},~\citealp{Blum21:Incentive,Toulis15:Design,Ashlagi14:Free,Hajaj15:Strategy}), but that direction is orthogonal to our paper. Studying the myriad of incentive issues in the allocation of other organs such as hearts is an important direction for future work. Recent work highlights widespread incentive misalignment in the US heart transplant allocation system and asserts that some of these issues could be mitigated with better allocation policies~\citep{Anagnostides26:Position}.

\paragraph{Imitation learning and learning-to-rank} Our methods presented in~\Cref{sec:learning} formalize potential function optimization as a form of offline imitation learning~\citep{Ross10:Efficient,Bain95:Framework}. We construct an omniscient oracle that optimally solves historical matching problems in hindsight to generate training labels. Most prior work on imitation learning has been studied in the context of \textit{Markov Decision Processes} for continuous control tasks, such as robotics, autonomous driving, or sensing~\citep{Argall09:Survey,Osa18:Algorithmic,Breitfeld25:Learning}. Its application to discrete online matching is understudied and is a shift from prior work in dynamic allocation, which largely relies on direct policy search from simulation outcomes~\citep{Dickerson12:Dynamic,Dickerson15:Futurematch,Anagnostides25:Policy}. The success of reinforcement learning in the organ allocation domain is limited by sparse rewards, high variance, and the computational cost of simulating patient trajectories~\citep{Yu21:Reinforcement}. By casting the problem as self-supervised learning from an oracle, we bypass these issues, achieving high sample efficiency and enabling the use of expressive, differentiable function approximators such as deep neural networks.

We formulate the potential function training as a form of learning-to-rank problem~\citep{Cao07:Learning,Burges10:Ranknet}. In our  formulation, the model is trained to minimize the divergence between its predicted rankings and the oracle's optimal node selection. Learning-to-rank is distinct from standard classification, which typically assumes fixed label sets or treats predictions independently. In organ allocation, the candidate set is dynamic and of variable size. For each donor, the model must identify the optimal patient against a changing pool of alternatives. Learning-to-rank objectives explicitly model these within-list comparisons. Learning-to-rank has proven successful across domains ranging from healthcare resource allocation~\citep{Kamran24:Learning} to ad auctions~\citep{Richardson07:Predicting}. Our work is the first to extend it to optimizing matching policies via potentials.

\paragraph{Online matching}

Online matching has seen great success in domains such as Internet advertising~\citep{Huang24:Online,Mehta07:Adwords,Richardson07:Predicting}. 
%
Seminal work in online matching has focused on maximizing competitive ratios under adversarial arrival sequences~\citep{Karp90:Optimal}. These algorithms provide strong theoretical guarantees, such as the optimal $1-1/e$ ratio. However, the assumption of an adversary is ill-suited for organ allocation, where patient and donor arrivals follow discernible biological and temporal patterns. It is worth pointing out that although dynamic matching involves an intrinsic tradeoff between short- and long-term value, there are conditions under which this tension is limited~\citep{Kerimov24:Dynamic}, and carefully designed greedy policies can be near-optimal~\citep{Kerimov25:Optimality}.

Our work is somewhat more closely aligned with the literature on online \textit{stochastic} matching, where arrivals are assumed to be drawn from known distributions~\citep{Feldman09:Online, Aggarwal11:Online}. This includes the edge-weighted model~\citep{Feldman09:Online}, which is analogous to an advertiser gaining specific revenue from a match, as well as the vertex-weighted model~\citep{Aggarwal11:Online}. The edge-weighted stochastic matching model was recently adapted to heart transplant allocation, showing that under certain realistic yet simplifying assumptions, theoretically near-optimal policies are possible~\citep{Zilberstein26:Near}. 
Data-driven approaches are commonly used to parameterize policies, such as in ad exchanges~\citep{Parkes05:Optimize} and dynamic kidney exchange~\citep{Awasthi09:Online,Dickerson12:Dynamic,Dickerson15:Futurematch}. However, standard stochastic matching algorithms typically assume that the underlying bipartite graph or type distribution is known \textit{ex ante}. These assumptions limit their applicability to the highly dynamic domain of organ allocation, where patients are continuously added to and removed from the waitlist due to medical progression, logistics, or death, creating a state space too complex for standard models. 

Another prominent direction in online matching leverages online primal-dual frameworks, where dual variables act as shadow prices to guide allocation. For example, dual mirror descent algorithms dynamically update thresholds to maximize efficiency~\citep{Balseiro20:Dual}. These approaches were extended to handle fairness constraints in regularized online allocation~\citep{Balseiro21:Regularized}. Parallel research is the emerging field of learning-augmented algorithms, which integrate machine-learned advice into classical frameworks~\citep{Antoniadis20:Secretary,Li23:Learning}. Finally, deep reinforcement learning has been applied to learn matching policies directly from experience~\citep{Alomrani22:Deep}. Our work differentiates itself from these approaches by adopting an imitation learning strategy: instead of relying on trial-and-error or online gradient updates, we explicitly learn to mimic a hindsight-optimal oracle to guide the online policy.

\section{Our heart transplant allocation model}
We model the heart transplant allocation process as a dynamic online bipartite matching problem. In this section, we define the model, including the dynamics, the utility, and the optimization objective.

The allocation process unfolds over a discrete horizon $\mathcal{H} = \{1, \dots, T\}$. At each time step $t$, a donor organ (online node) arrives. The decision maker must then decide whether to match the organ to a patient (offline node) in the waitlist or discard it. We refer to the sequence of events over $\mathcal{H}$ as a \textit{trajectory}. The system is defined by the following components:

\begin{itemize}
    \item $\mathcal{D}$: the stream of donors. At time $t$, a donor $d^{(t)} \in \mathcal{D}$ arrives, characterized by a vector of features that includes blood type, medical data, and location. We use the superscript notation $d^{(t)}$ to denote the donor arrival at time $t$.
    \item $\mathcal{P}^{(t)}$: the set of patients waiting for a transplant at time $t$. Each patient $p^{(t)} \in \mathcal{P}^{(t)}$ is defined by a vector of features including blood type, medical data, and location. The pool is dynamic: new patients arrive, and existing patients may depart (due to death or delisting) or experience medical status updates. The notation $p^{(t)}$ denotes a patient's condition at time $t$.
    
    \item $\mathcal{C}: \mathcal{D} \times \mathcal{P} \rightarrow  \{0,1\}$ is a binary constraint function. A match between donor $d$ and patient $p$ is feasible if and only if $\mathcal{C}(d, p) = 1$ (\textit{e.g.}, an edge exists in the bipartite graph). Constraints typically include blood type compatibility and maximum geographic distance.
    
    \item $\mathcal{U}: \mathcal{D} \times \mathcal{P} \rightarrow \mathbb{R}$: the utility function quantifying the value of a match. In this work, we maximize \textit{population life years gained (PLYG)}, defined formally below. While we focus on PLYG, the framework generalizes to arbitrary utility functions, including those that capture fairness or equity.
\end{itemize}

In the context of abstract online matching, $\mathcal{D}$ is the set of online nodes, $\mathcal{P}$ is the set of offline nodes (each with capacity $1$), $\mathcal{C}$ determines edge existence, and $\mathcal{U}$ determines edge weights. A key constraint in organ allocation is the viability of an organ. When a donor $d^{(t)}$ arrives, the allocation must be made immediately. An organ cannot be stored for future time steps. Therefore, the set of feasible matches for a donor $d^{(t)}$ is restricted to the set of compatible patients currently in the pool, denoted $\mathcal{P}_{d,t}$:
$$ \mathcal{P}_{d,t} = \left\{ p^{(t)} \in \mathcal{P}^{(t)} \mid \mathcal{C}\left(d^{(t)}, p^{(t)}\right) = 1 \right\}. $$
A policy must select a patient $p^{(t)} \in \mathcal{P}_{d,t}$ or discard the organ.

PLYG is a utilitarian metric for organ allocation, defined as the expected increase in survival time for a patient receiving a donor organ versus remaining unmatched on the waitlist. This metric is standard in organ allocation research (\emph{e.g.},~\citealp{Berrevoets20:OrganITE,Berrevoets21:Learning}). Maximizing PLYG incentivizes matches that provide the largest marginal gain in survival. PLYG does not simply target the sickest patients (who may have low post-transplant survival) or the healthiest (who have high waitlist survival), but rather takes a population-level efficiency-focused approach. Beyond PLYG, there are also other objectives one can consider for the application, including fairness with respect to different demographic groups. Our framework is agnostic to the underlying utility function. For example, the edge weights can be adjusted to prioritize marginalized populations, as has been done in kidney exchange research~\citep{Dickerson15:Futurematch}.
 
Let $X = \left\{\left(d^{(t)}, p^{(t)}\right)\right\}_{t=1}^T$ be a sequence of matches made over the horizon $\mathcal{H}$, where $t$ denotes the time of the match. The total utility is the sum of the marginal life years gained:
$$ \mathcal{U}(X) = \sum_{\left(d^{(t)}, p^{(t)}\right) \in X} \mathcal{U}\left(d^{(t)}, p^{(t)}\right). $$

In an offline setting, where the full sequence of donors and patient dynamics is known in advance, the optimal allocation $X^*$ is the maximum weight matching:
\begin{equation}\label{eq:optimal}
  X^* = \smash{\arg\max}_{X} \quad \mathcal{U}(X) \quad \text{s.t.} \quad \forall \left(d^{(t)}, p^{(t)}\right)\in X, \mathcal{C}\left(d^{(t)},p^{(t)}\right)=1.
\end{equation}

\begin{figure}
    \centering
    \begin{subfigure}[b]{0.45\linewidth}
        \centering
        \includegraphics[width=\linewidth]{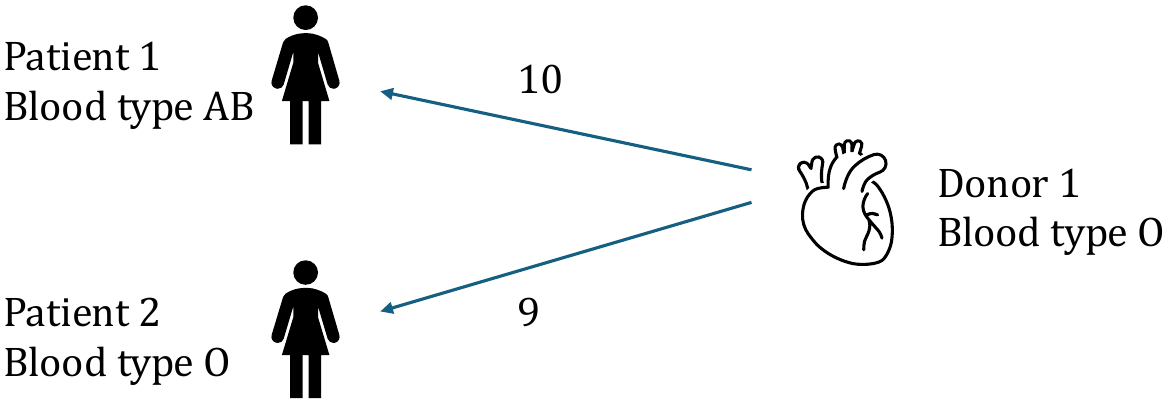}
        \caption{Matching at $t=1$.}
        \label{fig:ex1}
    \end{subfigure}
    \hfill
    \begin{subfigure}[b]{0.45\linewidth}
        \centering
    \includegraphics[width=\linewidth]{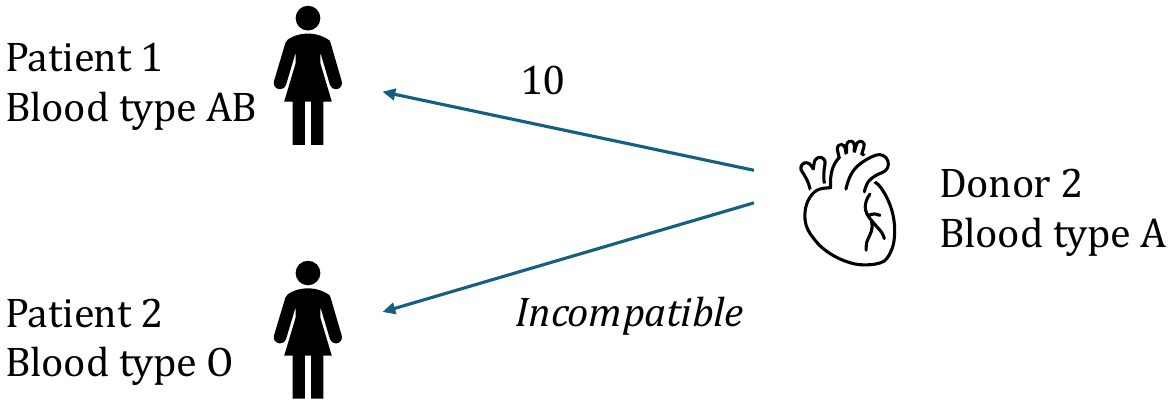}
        \caption{Matching at $t=2$.}
        \label{fig:ex2}
    \end{subfigure}   
    \caption{Example motivating non-myopic decision making in heart transplant allocation. The edge weights denote the immediate utility of a patient-donor match. A myopic algorithm would obtain a suboptimal utility of 10 by matching Patient 1 with Donor 1. The optimal decisions are non-myopic: match Patient 2 with Donor 1 and Patient 1 with Donor 2.}
    \label{fig:myopic-example}
\end{figure}

In the real-world online setting, the policy must select a match at time $t$ without knowledge of future donor arrivals or patient dynamics. We motivate the need for non-myopic decision making with a simple example illustrated in~\Cref{fig:myopic-example}. Consider a waitlist with two patients: Patient 1 with blood type AB and Patient 2 with blood type O. At $t=1$, a universal donor (\textit{i.e.}, blood type O) arrives. Matching to Patient 1 yields a slightly higher immediate utility than matching to Patient 2. A myopic policy would prefer Patient 1. However, at $t=2$, a type A donor arrives. Type A organs are compatible with Patient 1 but incompatible with Patient 2. If Patient 1 was matched at $t=1$, the type A organ must be discarded, resulting in a total utility of 10. In contrast, an optimal policy recognizes that Patient 1 is easier to match and allocates the type O organ for Patient 2. This sequence yields a utility of 19. In fact, the utility of Patient 2 can be an arbitrarily small positive number, and the optimal matching would still be the same. Preserving scarce resources for hard-to-match patients motivates our use of \textit{potential functions} to quantify the long-term value of patients on the waitlist.

\section{Potential-based matching}

In online matching, an algorithm must balance immediate gains against future uncertainty. Potential-based policies operate as greedy algorithms that act on learned, modified weights; \textit{potentials} are a data-driven approach for quantifying the future value of the resources being consumed~\citep{Dickerson12:Dynamic}. 


In the context of heart transplantation, considering the long-term consequences of matching patients on the waitlist carries high stakes. As seen in the motivating example above, allocating scarce organs to patients who may have many future matching opportunities can starve hard-to-match patients. Such inefficiency results in preventable waitlist mortality.

To explicitly quantify this future value in online matching problems, a learned \textit{potential function} augments the weights of edges to estimate the long-term strategic value of matching (or not matching) an offline node to an online node. We detail the potential-based policy framework in the context of heart transplantation.

\subsection{General policy framework}

We define a value function $Q\left(d^{(t)}, p^{(t)}, \mathcal{P}^{(t)}; \theta\right)$ that assigns a scalar value to every viable match between the current donor $d^{(t)}$ and a patient $p^{(t)}$ in the waitlist $\mathcal{P}_{d,t}$. This function, parameterized by $\theta$, augments the weights at each decision step in the matching problem. The weights $\theta$ are learned offline, while the online policy selects the patient that maximizes the score:
\begin{equation*}
    p_{selected} = \underset{p^{(t)} \in \mathcal{P}_{d,t}}{\arg\max} ~~Q\left(d^{(t)}, p^{(t)},\mathcal{P}^{(t)} ; \theta\right).
\end{equation*}

The augmented edge weight is defined as a combination of the actual edge weight and a parameterized potential function $P_\theta$:
\begin{equation*}
    Q\left(d^{(t)}, p^{(t)},\mathcal{P}^{(t)}; \theta\right) = \underbrace{\mathcal{U}\left(d^{(t)}, p^{(t)}\right)}_{\text{immediate utility}} - \underbrace{P_\theta\left(d^{(t)}, p^{(t)},\mathcal{P}^{(t)}\right)}_{\text{learned potential}}.
\end{equation*}

The potential function $P_\theta$ acts as a penalty for removing a patient from the waitlist (consuming a resource). 

The learning process, detailed later in~\Cref{sec:learning}, seeks to learn parameters $\theta$ such that the induced policy maximizes  global utility. A patient with low potential signals that matching immediately has high strategic value (\textit{e.g.}, a patient is in critical condition or harder to match). On the other hand, a high potential indicates that maintaining a patient in the pool is desired. Generalized to abstract online matching, the potential function can depend on the offline nodes, the arriving online node, and any associated features.

While the potential-based matching framework generalizes to arbitrary potential functions, in this work, we utilize linear functions and neural networks to bracket the spectrum of interpretability and expressiveness. Linear models provide a transparent baseline, allowing for the direct inspection of feature weights. Neural networks can capture complex interactions between patient and donor features and the state of the waitlist. Future work may investigate intermediate architectures, such as decision trees, which offer some non-linearity while retaining some interpretability.

\subsection{Linear potential functions}

The simplest form of potential function assumes that the potential of a patient is a linear combination of features:
\begin{equation*}
     P_\theta\left(d^{(t)}, p^{(t)},\mathcal{P}^{(t)}\right) = \theta^T \phi\left(d^{(t)}, p^{(t)},\mathcal{P}^{(t)}\right),
\end{equation*}
where $\phi$ is a feature mapping and $\theta$ is a learnable weight vector. Linear potentials offer high interpretability, but sacrifice expressiveness. All prior experiments on potential-based matching have relied exclusively on linear formulations~\citep{Dickerson12:Dynamic,Dickerson15:Futurematch,Anagnostides25:Policy}.

We utilize a feature set $\phi$ that encodes biological and geographic constraints, capturing two primary drivers of organ scarcity. Specifically, we use indicator variables for the four blood types (O, A, B, and AB) and the nine geographical regions of the mainland United States.

\subsection{Neural network-based potential functions}\label{sec:nn}

To capture non-linear interactions between patient characteristics and the state of the waitlist, we learn potentials using a neural network, specifically a \textit{Multi-Layer Perceptron (MLP)}. We design the MLP so that if $\theta = \Vec{0}$, the potential is zero; this effectively enables regularization to ground the model closer to a myopic policy and better learn subtle deviations.

In addition to a low-dimensional neural network that uses the patient's blood type as input, we employ a richer, 34-dimensional feature vector that captures both the characteristics of the specific match between the patient and donor and the broader context of the waitlist. Match features consist of a patient's blood type and location, the donor's blood type, and the predicted life years gained from the match. State features summarize the pool of waitlisted patients and the state of the allocation. These features include the fraction of waitlisted candidates with each blood type, the spatial distribution of the waitlist (fraction of patients in each region), the maximum and mean utility available in the current pool, and the donor arrival time. The input features are normalized using a standard scaler (zero mean, unit variance). Specific architectural details are provided in~\Cref{tab:hyperparams}.

\section{Learning the potentials}\label{sec:learning}

The objective of learning a potential-based policy is to find the parameter vector $\theta$ that maximizes the cumulative utility of an online matching over time. However, directly optimizing $\theta$ based on simulator outcomes poses significant computational challenges. The objective function is non-convex and non-differentiable with respect to $\theta$. The underlying system involves discrete, sequential decisions where small changes in weights can lead to discontinuities in the allocation policy.

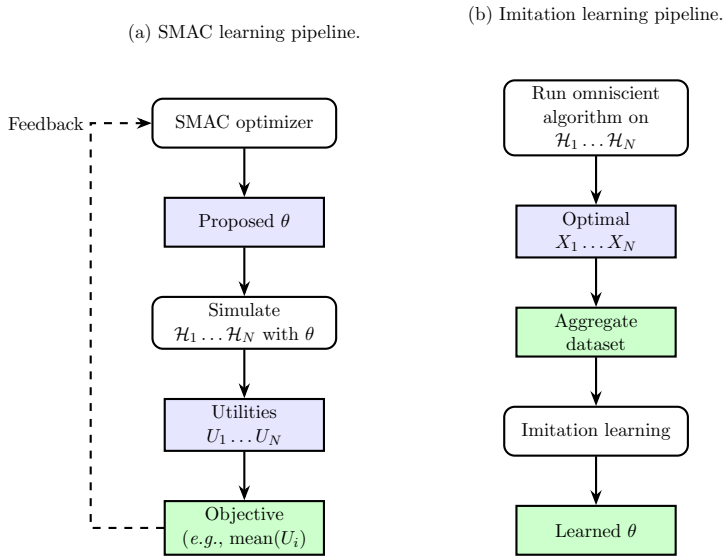
\begin{figure}[b!]
\centering
\scalebox{1.0}{\begin{tikzpicture}[
    scale=0.65, transform shape,
    node distance=1.0cm,
    block/.style={rectangle, draw, thick, rounded corners,
                  minimum height=1cm, text width=3.5cm, align=center},
    data/.style={rectangle, draw, thick, fill=blue!10,
                 minimum height=1cm, text width=3cm, align=center},
    agg/.style={rectangle, draw, thick, fill=green!20,
                minimum height=1cm, text width=3cm, align=center},
    arrow/.style={-{Stealth[length=2mm]}, thick},
    looparrow/.style={-{Stealth[length=2mm]}, thick, dashed}
]
\node (smac_title) {(a) SMAC learning pipeline.};
\node (smac) [block, below=of smac_title] {SMAC optimizer};
\node (theta) [data, below=of smac] {Proposed $\theta$};
\node (sim) [block, below=of theta] 
    {Simulate $\mathcal{H}_1 \dots \mathcal{H}_N$ with $\theta$};
\node (util) [data, below=of sim] {Utilities $U_1 \dots U_N$};
\node (obj) [agg, below=of util] {Objective (\textit{e.g.}, mean$(U_i)$};

\draw [arrow] (smac) -- (theta);
\draw [arrow] (theta) -- (sim);
\draw [arrow] (sim) -- (util);
\draw [arrow] (util) -- (obj);
\draw [looparrow] (obj.west) -- ++(-1.5,0) |- (smac.west) 
    node[midway, left] {Feedback};

\end{tikzpicture}
\hspace{1cm} 
\begin{tikzpicture}[
    scale=0.65, transform shape,
    node distance=1cm,
    block/.style={rectangle, draw, thick, rounded corners,
                  minimum height=1cm, text width=3.5cm, align=center},
    data/.style={rectangle, draw, thick, fill=blue!10,
                 minimum height=1cm, text width=3cm, align=center},
    agg/.style={rectangle, draw, thick, fill=green!20,
                minimum height=1cm, text width=3cm, align=center},
    arrow/.style={-{Stealth[length=2mm]}, thick},
    looparrow/.style={-{Stealth[length=2mm]}, thick, dashed}
]
\node (il_title) {(b) Imitation learning pipeline.};
\node (milp) [block, below=of il_title] 
    {Run omniscient algorithm on $\mathcal{H}_1 \dots \mathcal{H}_N$};
\node (decisions) [data, below=of milp] {Optimal \\ $X_1\dots X_N$};
\node (dataset) [agg, below=of decisions] {Aggregate dataset};
\node (il) [block, below=of dataset] {Imitation learning};
\node (theta_final) [data, fill=green!20, below=of il] {Learned $\theta$};

\draw [arrow] (milp) -- (decisions);
\draw [arrow] (decisions) -- (dataset);
\draw [arrow] (dataset) -- (il);
\draw [arrow] (il) -- (theta_final);


\end{tikzpicture}}
\caption{Learning pipelines using a set of trajectories. (a) SMAC uses the simulator as a black box, optimizing the average utility across trajectories in a feedback loop. (b) Imitation learning approaches first compute the optimal patient-donor allocations in the training set, $X_1,..., X_N$, using the omniscient algorithm for each trajectory, and then aggregate these decisions into a single dataset. The approaches then learn the potential function weights $\theta$ in a one-shot process.}
\label{fig:learning-pipelines}
\end{figure}

We explore two distinct optimization paradigms illustrated in~\Cref{fig:learning-pipelines}. The first approach treats the simulator as a black box and optimizes $\theta$ using Bayesian optimization, specifically, SMAC. The optimizer iteratively proposes a candidate $\theta$, runs the simulation, observes the total utility, and updates its probabilistic model of the response surface to propose the next $\theta$. Black-box optimization via SMAC is the standard approach utilized in all prior potential-based matching literature~\citep{Dickerson12:Dynamic,Anagnostides25:Policy,Dickerson15:Futurematch}.

The second---our primary novel contribution---is a self-supervised imitation learning framework for optimizing $\theta$. Instead of optimizing the simulation outcomes directly, the policy learns to mimic the decisions of an optimal omniscient algorithm that can make allocations in hindsight using full knowledge of events. In other words, the omniscient algorithm to be mimicked has perfect foresight. For a given donor, we train the potential function to reflect that the omniscient algorithm's selected patient was chosen over all other candidates in the pool. Across a large-scale dataset of decisions, the potentials reflect the logic used by the optimal matches, identifying in what contexts patients should be matched versus maintained in the pool. 

In the next two subsections we describe these two approaches, respectively.

\subsection{Black-box optimization}\label{sec:smac}

The black-box optimization approach has been used to optimize potentials in kidney exchange research~\citep{Dickerson12:Dynamic,Anagnostides25:Policy,Dickerson15:Futurematch}. As was done in that research, we will use SMAC~\citep{Hutter11:Sequential} for adjusting the potentials. SMAC treats the entire simulation and utility calculation as a black box function $f(\theta)\rightarrow \mathbb{R}$. It uses Bayesian optimization to model $f$ and iteratively selects a candidate $\theta$ to evaluate.

While SMAC optimizes the true objective directly, it suffers from severe sample inefficiency and has no finite-time convergence or quality guarantees. Each function evaluation requires simulating an entire trajectory. Furthermore, SMAC cannot scale to high-dimensional parameter spaces, such as the weights of a neural network. We use SMAC to benchmark the linear potential functions and show that it scales poorly in higher dimensions (\Cref{appendix:abblation-dist}). 

SMAC could also be used to minimize the loss functions of the imitation learning framework. However, SMAC is ill-suited for high-dimensional differentiable problems compared to gradient-based methods, which we will leverage in the next section.

\subsection{Our imitation learning framework}

Our imitation learning framework works using the optimal solutions from an omniscient algorithm, and we learn potential functions via supervised learning such that the induced ranking of patients mimics the optimal allocation. Our imitation learning framework offers significant advantages over black-box optimization. First, it is data-efficient. Obtaining ground-truth labels from the omniscient algorithm is computationally inexpensive and invariant to the dimensionality of the potential function. Second, it improves interpretability. Rather than optimizing a sparse, delayed reward signal, we explicitly train the policy to replicate specific decision patterns from the omniscient algorithm. Finally, the surrogate loss functions are differentiable (and convex in the linear case), which facilitates stable and efficient convergence using gradient-based optimization.

\subsubsection{Omniscient algorithm}

Supervising imitation learning requires ground-truth labels. We construct an omniscient algorithm that computes the optimal allocation over a fixed horizon, $\mathcal{H}$, in hindsight. The omniscient algorithm relies on perfect knowledge over the horizon, specifically, the complete sequence of donor arrivals and patient progressions. While using an omniscient algorithm to benchmark online algorithms is standard in operations research (\textit{e.g.}, for computing competitive ratios), directly learning from these algorithms is relatively unexplored. 

Let $\mathcal{D}$ be the set of donors arriving over a horizon $\mathcal{H}$ and $\mathcal{P}$ be the set of all unique patients who appeared on the waiting list at any time $t \in \mathcal{H}$. We construct a static bipartite graph where the nodes correspond to these patients and donors. An edge exists between a patient $p$ and a donor $d$ (having arrived at time $t$) if and only if they were compatible at that moment (\textit{i.e.}, $p\in \mathcal{P}_{d,t}$). The weight of the edge is the utility $\mathcal{U}\left(d^{(t)},p^{(t)}\right)$. The optimal allocation of transplants over $\mathcal{H}$ corresponds to a maximum weighted bipartite matching on this graph. 

We formulate the optimization as an integer linear program with binary variables as follows. Let $x_{d,p} \in \{0,1\}$ be a binary variable equal to $1$ if patient $p$ is matched with donor $d$, and $0$ otherwise. Note that variables exist only for compatible patient-donor pairs.

\begin{equation*}
\begin{array}{ll@{}ll}
\text{maximize}   & \displaystyle\sum_{d \in \mathcal{D}} \sum_{p \in \mathcal{P}_{d}} \mathcal{U}(d, p) \cdot x_{d,p} &\\
\text{subject to} & \displaystyle\sum_{p \in \mathcal{P}_{d}}   x_{d,p} \le 1,  &\forall d \in \mathcal{D}\\
                  & \displaystyle\sum_{d \in \mathcal{D}_{p}}   x_{d,p} \le 1,  &\forall p \in \mathcal{P}\\
                  & x_{d,p} \in \{0,1\} & \forall (d,p) \in \mathcal{E}
\end{array}
\end{equation*}
where $\mathcal{P}_d$ is the set of compatible patients for donor $d$, and $\mathcal{D}_p$ is the set of donors compatible with patient $p$, and $\mathcal{E}$ is the set of all viable matches. 

We refer to the above program as the \textit{omniscient algorithm}. The constraints enforce that each patient and donor is matched only once. The program outputs the optimal utility (according to~\Cref{eq:optimal}) over $\mathcal{H}$ as well as the optimal patient-donor pairs, $X^*$. We leverage the optimal patient-donor pairs, $X^*$, to construct a dataset of decisions for the imitation learning pipeline. The optimal utility also serves as a theoretical upper bound for evaluating online policies.

A typical instantiation of the integer program for a month-long horizon contains tens to hundreds of thousands of decision variables (corresponding to patient-donor matches). However, since the problem reduces to finding a maximum weight matching in a bipartite graph, the constraint matrix is \textit{totally unimodular}~\citep{Schrijver03:Combinatorial}. Total unimodularity guarantees that the optimal solution to the linear programming relaxation is integral. We can then relax the integer constraints and solve the problem as a continuous linear program in polynomial time. An instance of this size can be solved optimally in seconds using an off-the-shelf linear program solver (\textit{e.g.}, Gurobi). Combinatorial algorithms for maximum weight bipartite matching, such as the Hungarian algorithm, could alternatively be used to solve this problem.

\subsubsection{Learning potentials to imitate the omniscient algorithm}

We compile a dataset of optimal decisions from the omniscient algorithm consisting of tuples $\left(d^{(t)}, p^*,\mathcal{P}^{(t)}\right)$ for each $d^{(t)} \in \mathcal{D}$. The patient $p^*$ is the patient chosen by the omniscient algorithm from the waitlist $\mathcal{P}^{(t)}$ for donor $d^{(t)}$. We omit any donors from the training set that the omniscient algorithm discards.  For simplicity of presentation, we describe the approach as if we were learning from a single trajectory, though in practice we aggregate tuples from multiple trajectories.

Recall the potential-based policy:
\begin{equation*}
    Q\left(d^{(t)}, p^{(t)},\mathcal{P}^{(t)}; \theta\right) = \mathcal{U}\left(d^{(t)}, p^{(t)}\right) - P_\theta\left(d^{(t)}, p^{(t)},\mathcal{P}^{(t)}\right).
\end{equation*}

To imitate the omniscient algorithm, we require that the score of the optimal patient $p^*$ exceeds that of any other available candidate $p^{(t)}$:
\begin{equation}\label{eq:imitation-inequality}
    Q\left(d^{(t)}, p^{*},\mathcal{P}^{(t)}; \theta\right) > Q\left(d^{(t)}, p^{(t)},\mathcal{P}^{(t)}; \theta\right)
\end{equation} 
for all pairs $\left(p^*, p^{(t)}\right)$ in which $p^{(t)} \in \mathcal{P}_{d,t} \setminus \{p^*\}$. 

Using this inequality, we present two methods for learning potential functions. 

\vspace{.1in}
\noindent {\bf Support vector machine (SVM) formulation}

\noindent We first present a method for learning linear potential functions using a \textit{Support Vector Machine (SVM)} formulation. We frame the problem as finding the hyperplane that separates the optimal decision from suboptimal alternatives with a margin.

Assuming a linear potential function and enforcing a margin of $1$, we get from Inequality \ref{eq:imitation-inequality}:
\[ \mathcal{U}\left(d^{(t)}, p^{*}\right) - \theta^T \phi\left(d^{(t)}, p^{*},\mathcal{P}^{(t)}\right)> \mathcal{U}\left(d^{(t)}, p^{(t)}\right) - \theta^T \phi\left(d^{(t)}, p^{(t)},\mathcal{P}^{(t)}\right) +1 .\]

Rearranging terms:
\[  \theta^T \left[\phi\left(d^{(t)}, p^{(t)},\mathcal{P}^{(t)} \right) - \phi\left(d^{(t)}, p^{*},\mathcal{P}^{(t)} \right)\right] > 1 - \left[\mathcal{U}\left(d^{(t)},p^*\right) - \mathcal{U}\left(d^{(t)},p^{(t)}\right) \right] .\]

The target margin for the learnable component $\theta$ is dynamically adjusted based on the difference in utility between the optimal and suboptimal patient. The training vectors are the feature differences $\phi\left(d^{(t)}, p^{(t)},\mathcal{P}^{(t)}\right) - \phi\left(d^{(t)}, p^{*},\mathcal{P}^{(t)} \right)$.

We optimize $\theta$ by minimizing the pairwise hinge loss over all donors and compatible patients:
\begin{equation}\label{eq:svm}
        \mathcal{L}_{SVM}(\theta) = \sum_{d^{(t)}\in \mathcal{D}} \sum_{p^{(t)} \in \mathcal{P}_{d,t} \setminus \{p^*\}} \max\left\{0, 1 - \left[Q\left(d^{(t)}, p^{*},\mathcal{P}^{(t)}; \theta\right) - Q\left(d^{(t)}, p^{(t)},\mathcal{P}^{(t)}; \theta\right)\right]\right\}
\end{equation}
Minimizing this convex objective finds a hyperplane $\theta$ that maximally separates the optimal patient from suboptimal alternatives. Our  derivation is based on RankSVM~\citep{Joachims02:Optimizing}.

We can also incorporate weighting into the SVM. For example, getting a high-utility decision right may be more important than getting a low-utility decision right. 

\vspace{.1in}
\noindent {\bf Regression formulation}

\noindent We can alternatively formulate the learning problem from \Cref{eq:imitation-inequality} using regression-based methods~\citep{Cao07:Learning}. 

\paragraph{Pairwise regression} We maximize the likelihood that the optimal patient is ranked higher than any individual alternative. This is equivalent to binary cross-entropy on the score differences: 

\begin{equation}\label{eq:pair}  
\mathcal{L}_{pair}(\theta) =  \sum_{d^{(t)} \in \mathcal{D}} \sum_{p^{(t)} \in \mathcal{P}_{d,t} \setminus \{p^*\}} \log \left[1 + \exp\left(Q\left(d^{(t)}, p^{(t)},\mathcal{P}^{(t)}; \theta\right) - Q\left(d^{(t)}, p^{*},\mathcal{P}^{(t)}; \theta\right) \right) \right].
\end{equation} 

\paragraph{Listwise regression} Alternatively, listwise regression considers the entire list of patients simultaneously. Listwise regression is consistent with the Plackett-Luce model commonly used in recommendation systems~\citep{Luce59:Individual,Plackett75:Analysis}. We minimize the \textit{Kullback-Leibler (KL)} divergence between the predicted softmax distribution and the ground-truth distribution: 
    \begin{equation}\label{eq:list}
        \mathcal{L}_{list}(\theta) =   \sum_{d^{(t)} \in \mathcal{D}} -\log \left[ \frac{\exp\left(Q\left(d^{(t)}, p^{*},\mathcal{P}^{(t)}; \theta\right)\right)}{\sum_{p^{(t)} \in \mathcal{P}_{d,t}} \exp\left(Q\left(d^{(t)}, p^{(t)},\mathcal{P}^{(t)}; \theta\right) \right)} \right].
    \end{equation}

In the linear case, pairwise regression is analogous to \textit{logistic regression (LR)} on utility differences, while listwise regression is equivalent to multinomial logistic regression. 
The regression formulation allows us to extend the imitation learning framework to deep neural networks. Neural networks trade the direct interpretability of linear functions for greater expressiveness. We provide experiments of training with the different loss functions in~\Cref{appendix:abblation-nn}; it turned out that both loss functions are very effective, with the listwise objective performing slightly better with higher-dimensional neural networks.

\section{Generating sound semi-synthetic data for learning}\label{sec:semi-synthetic}

A major challenge in policy learning for organ allocation is data scarcity. The historical registry naturally contains only a single realized trajectory of events. Training strictly on a single historical realization risks overfitting. Furthermore, imitation learning is susceptible to \textit{covariate shift}: a policy can be inaccurate when it encounters parts of the state space not covered during training~\cite{Ross10:Efficient}.

To mitigate this, we employ data augmentation. We construct a diverse set of semi-synthetic trajectories by bootstrapping the historical data. The augmentation exposes the learner to a wider variety of trajectories of waitlist states and donor arrivals. We develop a method that ensures that the generated trajectories are statistically consistent with historical data in a medically meaningful way but differ in the relative timing of events so that multiple trajectories can be generated for training. 
Specifically, we want to maintain that in expectation, the following match the real historical data:
\begin{itemize}
\item the distribution of donor features,
\item the distribution of patient features, and 
\item the duration and trajectory of patients' disease progression on the waitlist.
\end{itemize}
Our algorithm for generating such semi-synthetic trajectories is shown in~\Cref{alg:semisynth}

\begin{algorithm}[!h]
\DontPrintSemicolon
\caption{Semi-synthetic trajectory generation}
\label{alg:semisynth}

\KwIn{Historical sets $\mathcal{D}_{\mathcal{H}}, \mathcal{P}_{\mathcal{H}}$ over horizon $\mathcal{H}$; New horizon $\mathcal{H}'$; Resampling bounds $D_{\min}, D_{\max}, P_{\min}, P_{\max}$.}
\KwOut{Semi-synthetic trajectory $\tau_{syn}$}

\BlankLine
Initialize empty trajectory $\tau_{syn}$\;
Sample volume $N_D \sim U(D_{\min}, D_{\max})$\;
Sample $N_D$ donors with replacement from $\mathcal{D}_{\mathcal{H}}$\;
\For{each sampled donor $d$}{
    Assign new arrival time $t \sim U(1, |\mathcal
    H'|)$\;
    Add $d^{(t)}$ to $\tau_{syn}$\;
}

Sample volume $N_P \sim U(P_{\min}, P_{\max})$\;
Sample $N_P$ patients with replacement from $\mathcal{P}_{\mathcal{H}}$\;
\For{each sampled patient $p$}{
    \eIf{$p \in \mathcal{P}_{0}$ (Waitlist initial set)}{
        Shift start time by offset $\delta \sim U(-k, k)$\ up to $t=0$\;
    }{
        Assign new arrival time $t \sim U(1, |\mathcal{H}'|)$\;
    }
    Relativize all patient events (medical progression, death) to new start time\;
    Add $p$ to $\tau_{syn}$\;
}
\Return{$\tau_{syn}$}\;
\end{algorithm}

From a single historical horizon $\mathcal{H}$, we generate a large set of semi-synthetic trajectories over sub-horizons $\mathcal{H}'$. For instance, if $\mathcal{H}$ represents a three-month-long trajectory, we generate trajectories over one month. Let $\mathcal{D}_{\mathcal{H}}$ and $\mathcal{P}_{\mathcal{H}}$ be the sets of all unique donors and patients, respectively, who were present during $\mathcal{H}$. Let $\mathcal{P}_{0} \subset \mathcal{P}_{\mathcal{H}}$ be the subset of patients already on the waitlist at the start of $\mathcal{H}$ (\textit{i.e.}, at $t=0$). We compute $D_{\min}$ and $D_{\max}$ as the minimum and maximum number of donors observed in any historical window of length $|\mathcal{H}'|$ over $\mathcal{H}$. Similarly, let $P_{\min}$ and $P_{\max}$ be the lower and upper bounds on the number of unique patients on the waitlist. For each synthetic trajectory, we sample the total volume of donors $N_D$ and patients $N_P$ uniformly from these bounds:
\begin{equation*}
N_D \sim {U}(D_{min}, D_{max}), \quad N_P \sim {U}(P_{min}, P_{max}).
\end{equation*}

We sample patients and donors uniformly with replacement from the historical population. We treat the real registry as a representative sample of the underlying clinical distribution, allowing us to generate an arbitrary number of independent training scenarios. 

It is also important to distinguish between the initial waitlist and new arrivals. For new arrivals, we assign a random arrival time $t \sim {U}(1, |\mathcal{H}'|)$. For patients on the waitlist, their trajectories are altered by a small offset $k$ up to $t=0$. This preserves the realistic structure of the waitlist while slightly altering the starting state across trajectories. In all cases, a patient's internal progression (\textit{e.g.}, time from arrival to death) is preserved relative to their new arrival time.  We include a brief analysis in~\Cref{appendix:semi-synthetic} to prove that the semi-synthetic trajectory generation preserves the key statistical properties of the underlying population.

\begin{figure}[t!]
\centering
\includegraphics[width=\textwidth]{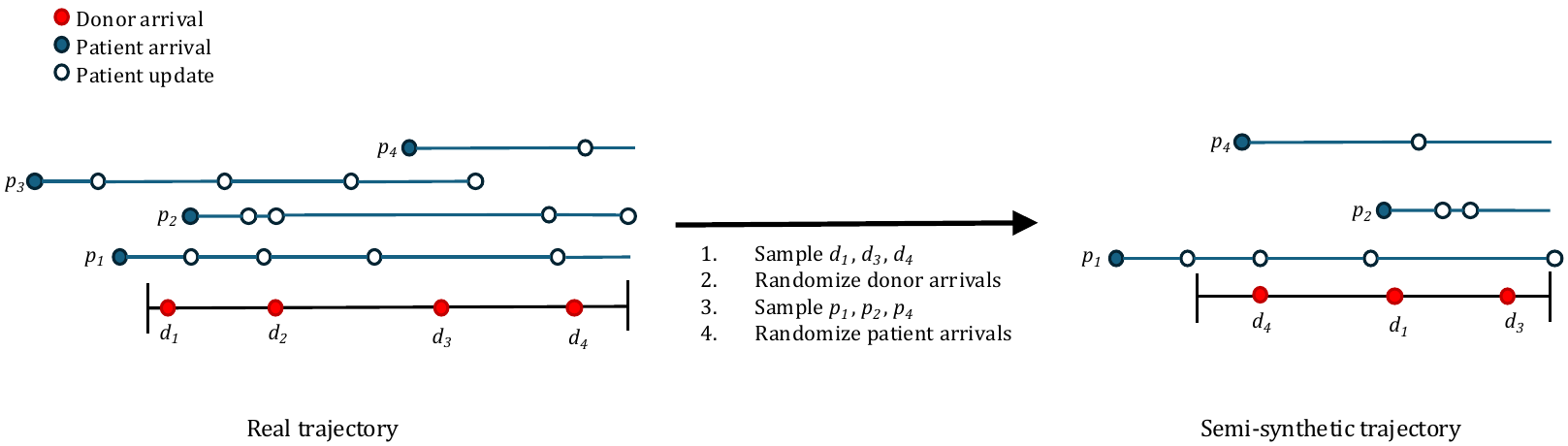}
\caption{Generation of a semi-synthetic trajectory from a single real trajectory. }
\label{fig:trajectory_transformation_visual}
\end{figure}

\Cref{fig:trajectory_transformation_visual} illustrates this process. Three patients and three donors are sampled from the real trajectory, and their introduction times are altered. However, the true patient progression is maintained. These new trajectories are semi-synthetic since the clinical features and progressions are real, but the relative timing and concurrent availability of patient-donor pairs are altered. The temporal shifts force learners to learn the underlying matching logic rather than memorizing historical timestamps. We show in~\Cref{appendix:abblation-semi-syn} that training with semi-synthetic data significantly improves the learned policies. 

\section{Baseline allocation policies for comparison}

We evaluate the potential-based policies against three allocation algorithms: the myopic utility maximizer, the current rule-based \textit{status quo} heart allocation policy in the US, and the proposed \textit{composite allocation score (CAS)}, the foundation of the continuous distribution system. We detail these algorithms below.

\subsection{Myopic algorithm}

The myopic algorithm serves as a baseline for greedy matching, maximizing the immediate life years gained without regard for future allocations. Formally, when a donor $d^{(t)}$ arrives, the algorithm selects:

\begin{equation*}
    p_{selected} = \underset{p^{(t)} \in \mathcal{P}_{d,t}}{\arg\max} ~~\mathcal{U}\left(d^{(t)}, p^{(t)}\right).
\end{equation*}

All policies---except the \textit{status quo} policy described next---will discard an organ if the utility of matching to $p_{selected}$ is less than or equal to zero (indicating a non-beneficial transplant) or the set of compatible patients is empty.  The myopic baseline explicitly optimizes the primary objective via greedy matching. However, as illustrated in~\Cref{fig:myopic-example}, it is prone to suboptimal decision making in dynamic environments.

\subsection{\textit{Status quo} in the United States}

The current allocation policy for adult heart transplants in the US is a hierarchical, rule-based system~\citep{Kilic21:Evolving}. Patients are placed into six mutually exclusive tiers based on medical urgency status, blood type match, and geographic proximity to the donor. The medical status is a measure of the severity of the patient's condition. It is determined by factors such as sensitization levels (CPRA) and mechanical circulatory support status (\textit{e.g.}, LVAD/ECMO). Patients within the same tier are sorted based on time on the waitlist. Priority is given to those who were listed earlier. Allocation follows a rigid lexicographic ordering. The \textit{status quo} policy is agnostic to value. It does not directly optimize post-transplant survival or population outcomes. It is primarily designed to reduce waitlist mortality among the sickest patients, often at the expense of overall system efficiency. A comprehensive presentation of the priorities is given in~\Cref{appendix:statusquo}.

\subsection{Continuous distribution and the composite allocation score (CAS)}

The continuous distribution framework, currently under review by the OPTN, proposes assigning a CAS to each candidate. CAS ranks patients using a linear combination of attributes:
\begin{equation*}
    p_{selected} = \underset{p^{(t)} \in \mathcal{P}_{d,t}}{\arg\max}  ~~\theta^\top \phi_{CAS}\left(p^{(t)}\right).
\end{equation*}
The feature vector $\phi_{CAS}$ includes characteristics such as medical urgency, sensitization (CPRA), waiting time, and blood type compatibility~\citep{Papalexopoulos24:Reshaping}.

Following public comment in 2024, the OPTN published preliminary patient features for CAS~\citep{OPTN2024:HeartUpdate}. However, as of early 2026, no weights have been proposed for the heart transplant CAS. To provide the strongest possible comparison, we optimize the weights for the CAS feature set using the potential learning framework from~\Cref{sec:learning}.  We treat CAS as a restricted policy class: a linear ranking function that does not explicitly observe the predicted life years gained. In~\Cref{appendix:abblation-cas} we show that learning weights for CAS via imitation learning substantially outperforms black-box optimization. 

There are two main key distinctions between CAS and our potential-based policies:
\begin{enumerate}
    \item CAS is a weighted linear sum of patient attribute values, whereas our method explicitly includes the immediate utility in decision making.
    \item CAS values are patient-centric (except for the geographic distance from the donor to the patient) and static relative to other patients. Our potential-based policies can include state features, allowing the policy to adapt to the composition of the waitlist.
\end{enumerate}

\Cref{tab:features} summarizes the information used by the different policies as of material published in early 2026. ``Explicit'' indicates that the feature is a direct component of the decision rule. ``Via utility'' indicates the feature is used implicitly to calculate the estimated survival $\mathcal{U}$. Only our potential-based policy explicitly accounts for the state of the waitlist and the direct utility simultaneously.
\begin{table}[!h]
    \centering
    \small
            \resizebox{\textwidth}{!}{
            \begin{tabular}{lcccc}
                \toprule
                 & \textbf{Status quo} & \textbf{CAS} & \textbf{Myopic} & \textbf{Potential-based (ours)} \\
                \midrule
                \textbf{Decision logic} & Hierarchical rules & Linear score & Greedy utility & Utility - potential \\
                \midrule
                \textit{Input features} & & & & \\
                Patient status & Explicit & Explicit & Via utility & Explicit \& via utility \\
                Blood type & Explicit & Explicit & Via utility & Explicit \& via utility \\
                Location  & Explicit & Explicit & Via utility & Explicit \& via utility \\
                Waiting time & Explicit & Explicit & Via utility & Explicit \& via utility \\
                Sensitization & Explicit & Explicit & Via utility & Explicit \& via utility \\
                Time on LVAD & Explicit & Explicit & Via utility & Explicit \& via utility \\
                Predicted LYG ($\mathcal{U})$ & --- & --- & Explicit & Explicit \\
                Waitlist state & --- & --- & --- & Explicit \\
                \bottomrule
            \end{tabular}
            }
        \caption{Comparison of information used in decision making across policies. ``Explicit'' indicates the feature is a direct component of the decision rule (\textit{e.g.}, a score component). ``Via utility'' indicates the feature is used implicitly to calculate the estimated survival $\mathcal{U}$. For the potential-based policies, all features can be explicit in the decision logic depending on the potential function.}
    \label{tab:features}
\end{table}

\section{Results}

We use a discrete-event simulator initialized with historical data from UNOS. The simulator models the complex dynamics of the allocation system, including donor arrivals, patient waitlist additions and removals (due to death or recovery), and the temporal progression of patient covariates (\textit{e.g.}, updating lab results).

We rely on Cox proportional hazards models~\citep{Cox72:Regression} to estimate the life years gained. Let $Y^T\left(d^{(t)}, p^{(t)}\right)$ denote the expected post-transplant survival (in years) for patient $p$ receiving organ $d$ at time $t$. Let $Y^W\left(p^{(t)}\right)$ denote the expected survival of patient $p$ at time $t$ if they do not receive a transplant and remain on the waitlist.  The utility of a match is defined as:

\begin{equation*}
  \mathcal{U}\left(d^{(t)}, p^{(t)}\right) \coloneq Y^T\left(d^{(t)}, p^{(t)}\right) - Y^W\left(p^{(t)}\right).
\end{equation*}

The predictors are trained on the UNOS patient registry containing transplant outcomes dating back to 1987. While more complex models have been developed for survival prediction~\citep{Lee18:Deephit}, their concordance is only marginally better than Cox regression. The simulator relies on these counterfactual predictors to compute patient outcomes. In fact, the simulator is agnostic to the survival model, allowing for modular upgrades. The simulator enforces compatibility constraints: a match is only feasible if the donor and patient have compatible blood types and are located within 1,000 nautical miles of each other. To ensure a uniform comparison, the distance constraint is applied to all policies, including the \textit{status quo}.

We employ a chronological split into a training horizon and a test horizon. We use the trajectory of events from the first quarter of 2019---January 1 to March 31---as the seed for training. As described in \Cref{sec:semi-synthetic}, we augment this period by generating semi-synthetic trajectories for model training. We evaluate the policies on the unseen, real trajectory of the remaining nine months (April 1 to December 31). We report metrics strictly on this period without data augmentation, ensuring that our results reflect performance on real, historical dynamics. For the omniscient algorithm, we use the Gurobi Optimizer version 12.0.3~\citep{Gurobi26:Gurobi}.

\subsection{Main comparison of approaches}

We compare the potential-based policies against the baselines, and benchmark them also against the omniscient algorithm.  We report the population life years gained (PLYG) of each policy in~\Cref{tab:real-lyg}. For potential-based policies and CAS, which rely on optimization, we select the best-performing variation from the tuning phase across feature sets and loss functions. For example, the CAS entry in~\Cref{tab:real-lyg} refers to the weights optimized via imitation learning with logistic regression, representing the strongest possible version of that framework. All ablations are presented in~\Cref{appendix:abblations}. \Cref{tab:real-lyg-sig} provides the statistical significance of these results.

\begin{table}[b!]
  \centering
  \resizebox{\textwidth}{!}{ 
  \begin{tabular}{l ccc ccccc c}
    \toprule
    & \multicolumn{3}{c}{Baselines} & \multicolumn{3}{c}{Linear pot.} & \multicolumn{2}{c}{Non-linear pot.} & \multicolumn{1}{c}{Upper bound} \\
    \cmidrule(lr){2-4} \cmidrule(lr){5-9} \cmidrule(lr){10-10}
    Month & Status quo & CAS & Myopic & SMAC & LR & SVM & NN (4) & NN (34) & Omniscient \\
    \midrule
    Apr. & 1332.56 & 2236.19 & 3119.34 & 3156.49 & 3123.28 & 3152.88 &\textbf{3161.46} & 3159.21 & 3328.70 \\
    May  & 1494.35 & 2503.84 & 3401.63 & 3413.65 & 3406.34 & 3423.36 & 3424.41 & \textbf{3426.39} & 3620.68 \\
    Jun. & 1335.29 & 2293.97 & 3176.31 & 3195.48 & 3186.39 & 3202.31 & 3195.55 & \textbf{3204.26} & 3383.43 \\
    Jul. & 1590.27 & 2470.58 & 3431.31 & 3412.50 & \textbf{3442.54} & 3441.79 & 3428.72 & 3441.19 & 3621.28 \\
    Aug. & 954.02  & 2607.67 & 3707.53 & 3742.73 & 3708.29 & 3724.13 & \textbf{3747.19} & 3744.17 & 3937.68 \\
    Sep. & 1632.99 & 2428.52 & 3380.64 & 3382.23 & 3390.62 & 3388.08 & 3389.69 & \textbf{3397.23} & 3590.33 \\
    Oct. & 1772.68 & 2341.80 & 3184.58 & 3203.56 & 3186.07 & \textbf{3209.87} & 3208.22 & 3204.37 & 3377.67 \\
    Nov. & 1705.20 & 2103.34 & 2867.00 & 2883.59 & 2864.91 & 2881.16 & 2882.68 & \textbf{2886.60} & 3011.73 \\
    Dec. & 983.93  & 2198.27 & 3253.35 & 3237.31 & 3249.54 & 3247.10 & 3245.60 & \textbf{3255.66} & 3404.93 \\
    \midrule 
    Mean PLYG  & 1422.37 & 2353.80 & 3280.19 & 3291.95 & 3284.22 & 3296.74 & 3298.17 & \textbf{3302.12} & 3475.16 \\
    \bottomrule
  \end{tabular}}
  \caption{Monthly and mean PLYG of policies on real trajectories from April to December 2019. LR $=$ logistic regression, SVM $=$ support vector machine, NN (4) $=$ 4-feature neural network, and NN (34) $=$ 34-feature neural network.}
  \label{tab:real-lyg}
\end{table}

We observe that CAS offers a substantial improvement over the \textit{status quo} policy, achieving an average of nearly 1,000 additional life years per month. CAS offers the flexibility to assign a granular score coupled with objective-aware weight optimization. Together, these features yield a far more efficient policy than the \textit{status quo}. However, our myopic baseline achieves a further improvement of $\sim$900 life years per month over CAS. These findings confirm that current allocation rules, and even proposed scoring systems like CAS, rely on features that are only proxies for survival outcomes. Explicitly incorporating survival predictions is an important driving factor for improving population outcomes.

The potential-based methods consistently further outperform the myopic baseline with statistical significance. Beyond SMAC, the imitation learning framework facilitates higher-utility allocations. Employing neural network potential functions achieves the strongest performance in 7 out of 9 test months, resulting in the highest mean utility. While the immediate utility drives the majority of the gain, the learned potentials successfully yield an additional 22 life years per month compared to greedy matching. The difference is equivalent to dozens of additional successful transplants per year.

We also observe a hierarchy in model performance. The neural network potential functions outperform the linear models. Furthermore, the 34-feature neural network outperforms the 4-feature variant. These results suggest that the state of the waitlist provides an important signal that allows the policy to adapt to the allocation dynamics.

Finally, our best potential-based model achieves a competitive ratio of around 95\% compared to the theoretical upper bound from omniscient matching, making our approach near-optimal on the real data distribution. It is unknown whether closing the remaining 5\% gap is possible without knowledge of the future.

\subsection{Fairness and other secondary objectives}

\begin{figure}[b!]
    \centering
    \begin{subfigure}[b]{0.49\linewidth}
        \centering
        \includegraphics[width=\linewidth]{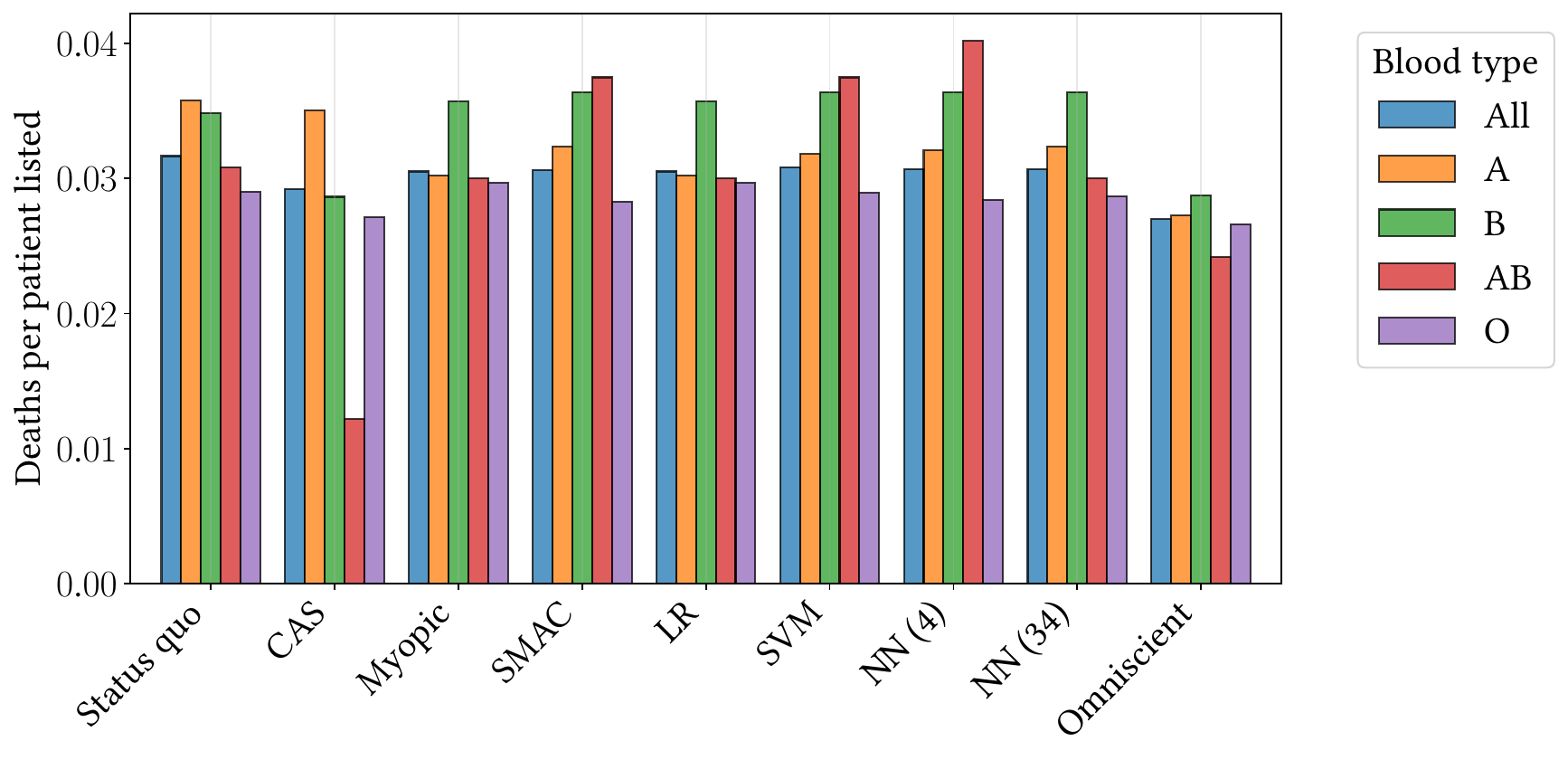}
        \caption{Mortality rate per patient.}
        \label{fig:death-rates-patient}
    \end{subfigure}
    \hfill
    \begin{subfigure}[b]{0.49\linewidth}
        \centering
    \includegraphics[width=\linewidth]{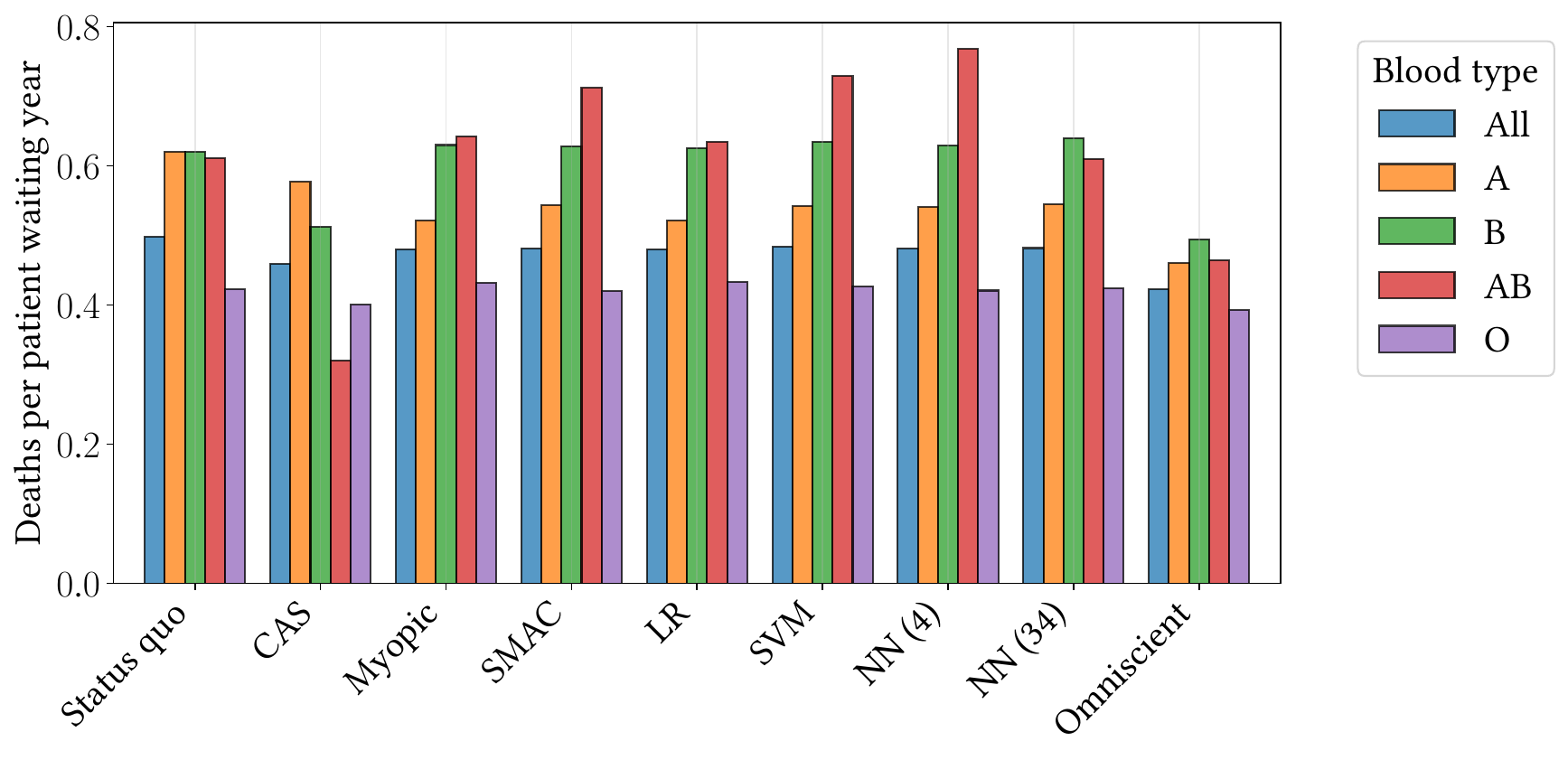}
        \caption{Mortality rate per wait year.}
        \label{fig:death-rates-waityear}
    \end{subfigure}   
    \caption{Average waitlist mortality rates normalized by number of patients in each group (a) and cumulative years on the waiting list per group (b).}
    \label{fig:death-rates}
\end{figure}

\begin{figure}[b!]
    \centering
    \begin{subfigure}[b]{0.49\linewidth}
        \centering
        \includegraphics[width=\linewidth]{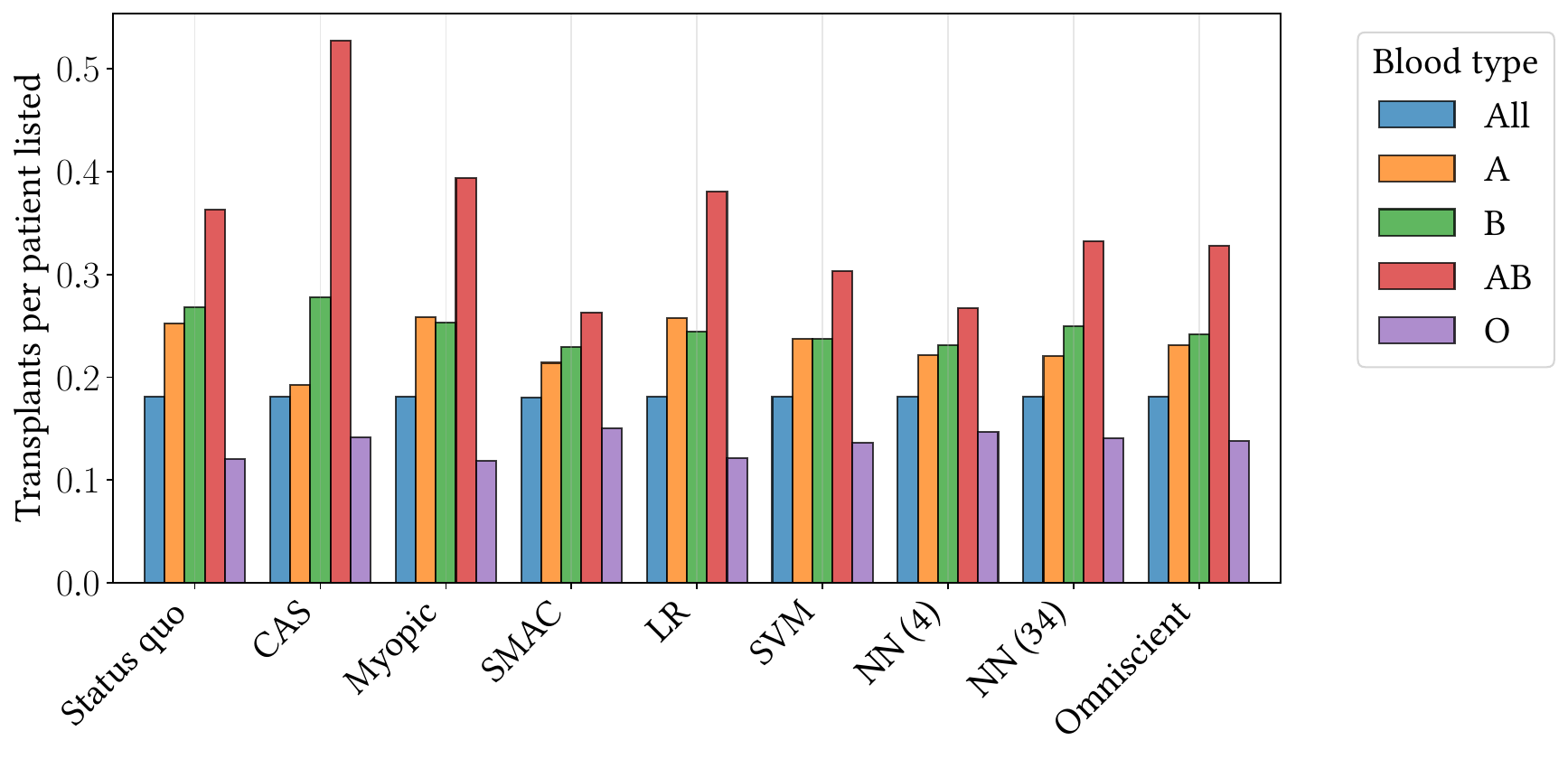}
        \caption{Transplant rate per patient.}
        \label{fig:transplant-rates-patient}
    \end{subfigure}
    \hfill
    \begin{subfigure}[b]{0.49\linewidth}
        \centering
    \includegraphics[width=\linewidth]{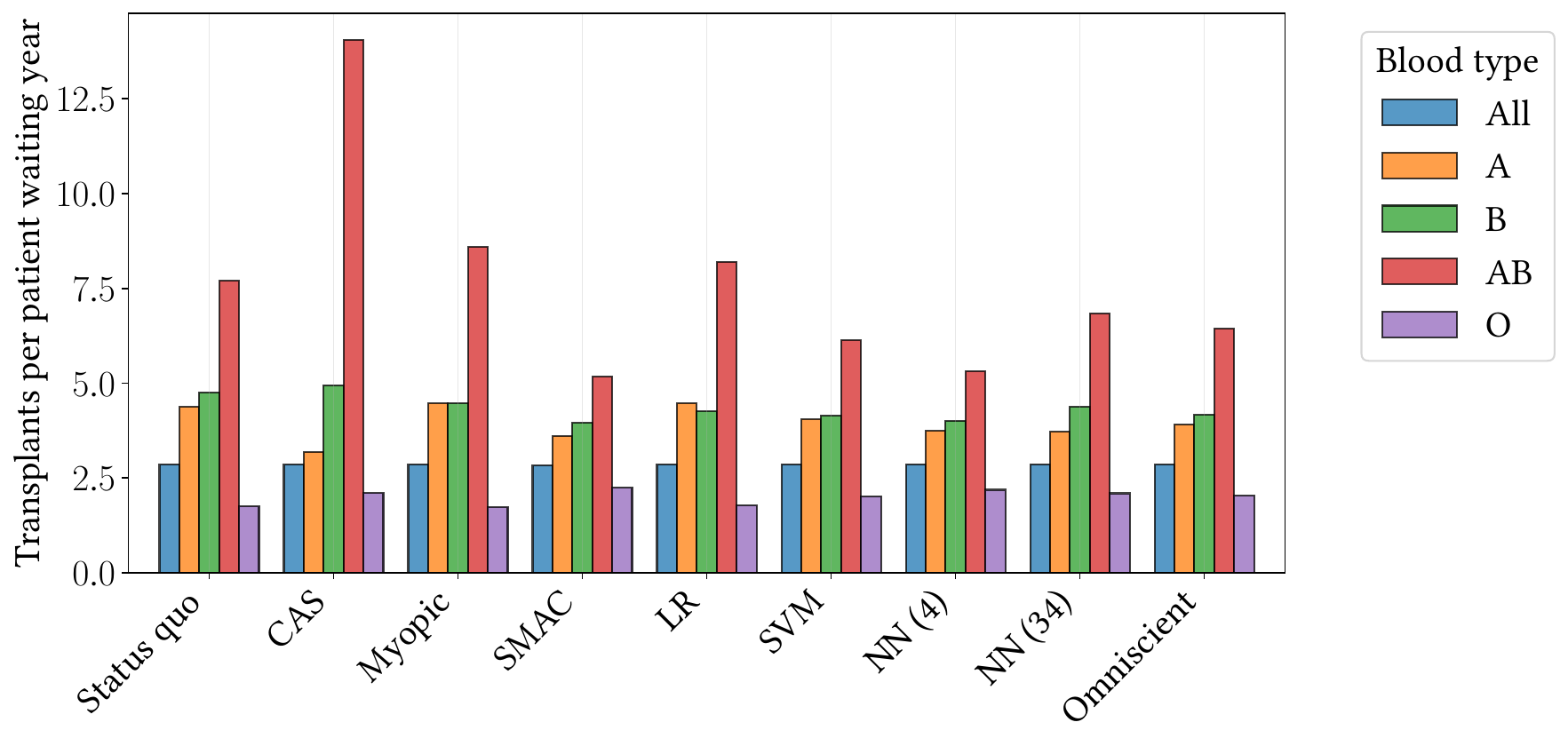}
        \caption{Transplant rate per wait year.}
        \label{fig:transplant-rates-waityear}
    \end{subfigure}   
    \caption{Average transplant rates normalized by number of patients in each group (a) and cumulative years on the waiting list per group (b).}
    \label{fig:transplant-rates}
\end{figure}

We also report important secondary objectives, specifically mortality rates and transplant rates across blood types, in~\Cref{fig:death-rates,fig:transplant-rates}. For both metrics, we report rates normalized by the number of patients listed and the cumulative time spent waiting.


The policies compare similarly across the aggregate patient population. Interestingly, the omniscient algorithm yields the lowest mortality rate, indicating that maximizing PLYG correlates with reducing waitlist mortality. CAS successfully lowers mortality rates compared to the \textit{status quo}, a primary goal of the continuous distribution system.

However, when disaggregated by blood type, significant disparities appear. Across all policies, patients with blood type AB experience the highest transplant rates and the highest number of transplants per waiting year. The advantage is most pronounced under CAS, in which AB transplant rates spike and mortality rates drop disproportionately. Blood type O patients consistently face the lowest transplant rates and the highest mortality risks. The disparity is partly structural. Type O patients constitute the largest group on the waitlist but can only receive type O organs, creating more competition for compatible donors.

The analysis of deaths per patient waiting year highlights the trade-offs in allocation. While CAS minimizes mortality for blood type AB, it does so at the cost of higher death rates for types A and B. Potential-based policies appear to smooth the extreme variances seen in CAS, resulting in a more equitable distribution of transplant rates among blood types A, B, and AB. The potential-based policies achieve higher utility while simultaneously sacrificing less fairness. Yet even potential-based policies suffer from systemic disparities in access across different blood types. Using our methodology, designing potentials that explicitly penalize inequality is a feasible direction for future work.

\section{Limitations}
\label{sec:limitations}

As algorithmic decision-making is increasingly used in healthcare, it is important to highlight the limitations of these approaches. First, we rely on historical data from the UNOS registry, which is imperfect. For example, it contains entry errors and missing values. While we implement preprocessing steps to impute missing entries and filter incongruent records, inherent data quality issues remain. In turn, the edge weights in our bipartite graph rely on imperfect predictive models. Determining patient outcomes, such as graft survival after a transplant, is a challenging predictive task. Unobserved confounders may influence actual outcomes. The historical registry also contains systemic bias that could be translated into the predictions. That being said, our potential learning framework is agnostic to how the weights are set. Therefore, future improvements in survival prediction and bias mitigation can be immediately combined with our algorithms. 

We optimized our policies to maximize the population life years gained, one of many possible objectives for the application. We evaluate policies against a multi-objective landscape. However, other objectives could also be optimized for, and we do not necessarily argue that utilitarian maximization is the sole or most appropriate metric for organ allocation. 

Finally, the experiments in this paper assumed that all offers are accepted by transplant centers. In reality, transplant centers can refuse matches. Our approach can naturally accommodate probabilistic edge existence by interpreting edge weights as expected values, which is consistent with the online stochastic matching model.

\section{Conclusions and future research}

In this work, we presented a computational framework for non-myopic policy optimization for dynamic matching problems based on potentials. We addressed the challenge of balancing immediate utility with long-term value in a dynamic environment. Our methodology represents a shift from prior work on optimizing potentials, moving from inefficient black-box optimization to a scalable offline imitation learning approach. Our framework enabled us to train expressive high-dimensional potential functions, a capability that prior optimization methods could not achieve.

We applied our framework to heart transplant allocation, a complex and high-stakes online matching problem. By leveraging our approach to learn potentials defined by neural networks, we optimized expressive policies that were previously infeasible. Our policies demonstrated significant improvements in population-level outcomes compared to myopic strategies, current and proposed US allocation systems, and prior potential-based baselines. Our policies are also near-optimal: they achieve 95\% of the solution quality upper bound that comes from omniscient matching. 

Future work can build on our framework. One can learn more interpretable potential functions, such as decision trees, or higher-dimensional functions, including deeper neural networks acting on a richer feature set. 

As the field of organ transplantation moves toward more flexible allocation models, the ability to optimize for long-term outcomes becomes increasingly vital. However, our analysis highlighted that high-utility policies do not automatically guarantee fair outcomes. Organ allocation is inherently multi-objective, and future work could explore policies that explicitly balance efficiency with equity. Our methodology enables such studies. This could take many directions. One could pre-process edge weights to give more weight to disadvantaged patient segments. Or, one could modify the omniscient algorithm to perform constrained optimization, generating potential functions that mimic a fair oracle.

Our analysis and methods are particularly timely as US policymakers review the implementation of continuous distribution for heart transplant allocation. Our findings suggested that the current linear scoring rules are overly restrictive and that substantial gains in population outcomes are possible through the more expressive, data-driven policies we proposed. Ultimately, we provided a scalable foundation for designing allocation mechanisms that make the most efficient use of scarce life-saving resources.

\section*{Acknowledgments}

Tuomas Sandholm and his PhD students Ioannis Anagnostides and Itai Zilberstein are supported by NIH award A240108S001, the Vannevar Bush Faculty Fellowship ONR N00014-23-1-2876, and National Science Foundation grant RI-2312342. Itai Zilberstein is also supported by the NSF Graduate Research Fellowship Program under grant DGE2140739. Arman Kilic is supported by NIH RO1 grant 5R01HL162882-03 which contributed to the funding for completion of this project. Arman Kilic is a speaker and consultant for Abiomed, Abbott, 3ive, and LivaNova, and founder and owner of QImetrix. All additional authors have no financial relationships to disclose. Any opinions, findings, and conclusions or recommendations expressed in this material are those of the author(s) and do not necessarily reflect the views of the funding agencies. We are indebted to Carlos Martinez from UNOS for answering numerous questions.

\bibliography{references}

\clearpage
\appendix

\setcounter{table}{0}
\renewcommand{\thetable}{A\arabic{table}}

\setcounter{figure}{0}
\renewcommand{\thefigure}{A\arabic{figure}}

\section{Properties of semi-synthetic trajectories}
\label{appendix:semi-synthetic}

We provide a brief analysis to demonstrate that the semi-synthetic trajectory generation preserves the key statistical properties of the underlying population.

\begin{proposition}[Volume consistency]
Let $D'$ and $P'$ be the random variables for the number of donors and patients in a semi-synthetic trajectory. The expected volumes are the midpoints of the observed historical range:
\begin{equation}
    \mathbb{E}[D'] = \frac{D_{\min} + D_{\max}}{2}, \quad \mathbb{E}[P'] = \frac{P_{\min} + P_{\max}}{2}.
\end{equation}
\end{proposition}

The sampling explores the full range of observed donors and patients, ensuring the semi-synthetic trajectories capture both high-congestion and low-resource scenarios.

\begin{proposition}[Feature invariance]
Let $\phi_{p}'$ be the feature vector of a randomly sampled patient in the synthetic trajectory (\textit{e.g.}, blood type, urgency, location). Let $\bar{\phi}_{\mathcal{H}}$ be the mean feature vector of the historical population $\mathcal{P}_{\mathcal{H}}$. Because patients are sampled uniformly at random, the expected feature vector satisfies:
\begin{equation}
    \mathbb{E}[\phi_{p}'] = \bar{\phi}_{\mathcal{H}}.
\end{equation}
\end{proposition}

The semi-synthetic generation process randomizes the temporal aspect of patient arrivals, but preserves the representative distribution of patient features, including blood types and medical urgency.

\begin{proposition}[Waitlist structure preservation]
Let $P'_0$ be the random variable for the size of the initial waitlist at the start of $\mathcal{H}'$. The expected value of $P'_0$ is:
\begin{equation}
    \mathbb{E}[P'_0] = \mathbb{E}[P'] \cdot \left( \frac{|\mathcal{P}_{0}|}{|\mathcal{P}_{\mathcal{H}}|} \right).
\end{equation}
\end{proposition}

The ratio of pre-existing patients to new arrivals remains consistent, maintaining a realistic waitlist composition.

\section{Ablation studies}
\label{appendix:abblations}

We compare the potential-based policies across two axes: the complexity of the potential function (linear vs. neural network and dimensionality of input) and the optimization method (black-box vs. imitation learning).

We evaluate linear models  with varying feature granularity, including a 4-dimensional feature vector encoding patient blood type and a 13-dimensional variation encoding blood type and location. These models are trained using three distinct optimizers: i) SMAC, ii) support vector machines (SVM), and iii) logistic regression (LR). For SMAC, we allocate a budget of 100 iterations. We observe that SMAC typically converges after approximately 50 iterations, suggesting it found a local maximum. Executing SMAC for this many iterations takes orders of magnitude more time compared to the imitation learning, which runs in minutes. 

We evaluate two  neural network  architectures: a small network using only patient blood type as input (4 features), and a larger network using a 34-dimensional state vector detailed in~\Cref{sec:nn}. The neural networks are trained exclusively via the imitation learning framework (using listwise and pairwise regression), as SMAC is ill-suited for high-dimensional parameter spaces. The hyperparameters of the neural networks are detailed in~\Cref{tab:hyperparams}.

\subsection{Learning with semi-synthetic data}
\label{appendix:abblation-semi-syn}

We evaluate the impact of data augmentation by comparing policies learned on the single real trajectory (January--March 2019) versus those trained on semi-synthetic variations. \Cref{tab:real-diff} reports the average difference in monthly utility between the augmented and non-augmented versions. A positive difference indicates that the semi-synthetic training yielded a higher utility. Across almost all policies, training on semi-synthetic data provides a substantial improvement. This benefit is most pronounced for CAS and the 34-feature neural networks, confirming that data augmentation effectively mitigates overfitting as the dimensionality and expressiveness of the policy increase.

\begin{table}[h!]
  \centering
  \small
  \begin{tabular}{lllr}
    \toprule
    Policy & Input dimension & Optimization method & Synthetic vs. real \\
    \midrule
    CAS & 14 & SMAC & +135.58 \\
        & 14 & SVM & -30.99 \\
        & 14 & LR & +4.91\\
    
    \addlinespace
    Linear potential & 4 & SMAC & +4.14 \\
                     &  4 & SVM & +4.94 \\
                     &  4 & LR & +4.23 \\
    \addlinespace
    Neural network potential & 4  & Pairwise loss & +3.01 \\
                   &   4 & Listwise loss & +3.08 \\
    \addlinespace
      & 34 & Pairwise loss & +41.30 \\
                   &   34 & Listwise loss & +9.91 \\
    \bottomrule
  \end{tabular}
  \caption{Difference in mean PLYG (utility) between algorithms trained on real versus semi-synthetic data. A positive difference indicates the semi-synthetic version performs better.}
  \label{tab:real-diff}
\end{table}

\subsection{Optimizing composite allocation score}
\label{appendix:abblation-cas}

We explore using both imitation learning and black-box optimization to learn the weights of CAS. We summarize the results in~\Cref{tab:cas}. As before, we learn the weights using the semi-synthetic trajectories from January-March 2019 and evaluate on April-December 2019. Using the imitation learning framework with both SVM and LR significantly outperforms SMAC. These results highlight the scalability limitations of black-box optimizers in high-dimensional spaces. We also see that the imitation learning framework is effective at learning general policies beyond potential-based ones. 

\begin{table}[h!]
  \centering
  \small
  \begin{tabular}{lccc}
    \toprule
    & \multicolumn{1}{c}{Black-box} & \multicolumn{2}{c}{Imitation learning} \\
    \cmidrule(lr){2-2} \cmidrule(lr){3-4}
    & SMAC & SVM & LR \\
    \midrule
    Mean PLYG & 1688.22 & 2322.30 & \textbf{2353.80} \\
    \bottomrule
  \end{tabular}
  \caption{Comparison of mean PLYG per month (utility) between CAS with weights learned via black-box optimization and imitation learning methods.}
  \label{tab:cas}
\end{table}

\subsection{Distance in linear potential functions}
\label{appendix:abblation-dist}

\Cref{tab:distance} compares linear potential-based policies using the full 13-dimensional feature set (including location) versus a restricted 4-dimensional set (blood type only). Increasing model complexity does not universally improve performance; for SMAC and SVM, the additional dimensions degrade performance. SMAC significantly struggles as the search space expands. LR is the only method to successfully leverage the expanded feature set. These results suggest that patient location may act as a noisy signal in linear models, complicating optimization without providing a strong predictive signal for allocation efficiency.

\begin{table}[h!]
  \centering
  \small
  \begin{tabular}{lccc} 
    \toprule
    & \multicolumn{1}{c}{Black-box} & \multicolumn{2}{c}{Imitation learning} \\
    \cmidrule(lr){2-2} \cmidrule(lr){3-4}
    & SMAC & SVM & LR \\
    \midrule
    Difference in PLYG & -11.93 & -6.82 & +3.96 \\
    \bottomrule
  \end{tabular}
  \caption{Difference in mean PLYG per month between linear policies with 13 dimensions (location \& blood type) versus 4 dimensions (blood type only). A positive difference favors the 13-dimensional model.}
  \label{tab:distance}
\end{table}

\subsection{Neural network loss functions}
\label{appendix:abblation-nn}

\Cref{tab:losses} compares the performance of neural network policies trained with pairwise (\Cref{eq:pair}) versus listwise (\Cref{eq:list}) regression losses. The 34-feature models generally outperform the 4-feature configurations. We hypothesize that the pairwise loss benefits the low-dimensional model by providing a denser signal through $O(N^2)$ comparisons, stabilizing learning when feature information is scarce. In contrast, the high-dimensional model has sufficient capacity to exploit the listwise loss, which directly optimizes the global ranking distribution.

\begin{table}[h!]
  \centering
  \small
  \begin{tabular}{lrrrr}
    \toprule
    & \multicolumn{2}{c}{4 features} & \multicolumn{2}{c}{34 features} \\
    \cmidrule(lr){2-3} \cmidrule(lr){4-5}
    Loss function & Pairwise & Listwise & Pairwise & Listwise \\
    \midrule
    Mean PLYG & 3298.17 & 3291.86 & 3299.97 & 3302.12 \\
    \bottomrule
  \end{tabular}
  \caption{Comparison of mean PLYG per month (utility) between potential-based policies using neural networks trained with weights learned via pairwise vs listwise loss.}
  \label{tab:losses}
\end{table}

\section{Omitted tables and figures}
\label{appendix:tables}

\begin{table}[hbtp!]
    \centering
    \small
    \begin{tabular}{llc}
    \toprule
    \textbf{Category} & \textbf{Hyperparameter} & \textbf{Value} \\
    \midrule
    \textbf{Architecture} & Hidden layers (4 features) & (64, 32) \\
     & Hidden layers (34 features) & (128, 64, 32) \\
     & Activation function & Leaky ReLU \\
     & Leak coefficient & $1 \times 10^{-2}$ \\
     & Output activation & Identity \\
    \midrule
    \textbf{Training} & Optimizer & Adam \\
     & Learning rate & $1 \times 10^{-4}$ \\
     & Batch size & 32 \\
     & Epochs & 25 \\
     & Gradient clipping (norm) & 1.0 \\
     & L2 regularization & $1 \times 10^{-3}$ \\
     & Dropout rate & 0.3 \\
    \bottomrule
    \end{tabular}
    \caption{Hyperparameters for neural network potential functions.}
    \label{tab:hyperparams}
\end{table}

\begin{table}[hbtp!]
\centering
\small
\begin{tabular}{l|ccccccccc}
\toprule
 & Status quo & CAS & Myopic & SMAC & LR & SVM & NN (4)  & NN (34) & Omniscient \\
\midrule
Status quo & -- & \checkmark & \checkmark & \checkmark & \checkmark & \checkmark & \checkmark & \checkmark & \checkmark\\
CAS &  & --  & \checkmark & \checkmark & \checkmark & \checkmark & \checkmark & \checkmark & \checkmark \\
Myopic &  &  & -- & $\star$ & $\star$ & \checkmark & \checkmark & \checkmark & \checkmark \\
SMAC &  &  &  &  -- &  &  & \checkmark & \checkmark & \checkmark \\
LR &  &  &  &  & -- & $\star$ & $\star$ & \checkmark & \checkmark \\
SVM &  &  &  &  &  & -- &  & $\star$ & \checkmark \\
NN (4) &  &  &  &  &  &  & -- & & \checkmark \\
NN (34) &  &  &  &  &  &  &  & -- & \checkmark \\
\bottomrule
\end{tabular}
\caption{Statistical significance of Wilcoxon paired signed-rank test for Table \ref{tab:real-lyg}. Significance is based on monthly outcomes. A $\checkmark$ denotes statistical significance ($p < 0.05$) and $\star$ denotes a weaker significance ($ 0.05 <p < 0.10$).}
\label{tab:real-lyg-sig}
\end{table}

\newpage
\section{\emph{Status quo} heart transplant allocation policy}
\label{appendix:statusquo}

The US heart transplant allocation policy is a hierarchical system that prioritizes primarily based medical urgency status, blood type compatibility, and geographic proximity. It comprises 68 priority tiers, where 1 represents the highest priority. Each tier is defined by a combination of i) medical status ranging from status 1 (highest urgency) to status 6; ii) blood type compatibility (where \emph{secondary} blood compatibility applies only between type O donors and type A or type AB patients); and geographic proximity between donor and recipient location. \Cref{tab:priority_tiers} contains the detailed description of each tier.

\begin{table}[htbp]
\footnotesize
\centering
\caption{\emph{Status quo} priority tiers.}
\label{tab:priority_tiers}
\begin{tabular}{clll@{\hspace{1em}}|@{\hspace{1em}}clll}
\toprule
\textbf{Tier} & \textbf{Status} & \textbf{Blood match} & \textbf{Distance (nm)} & \textbf{Tier} & \textbf{Status} & \textbf{Blood match} & \textbf{Distance (nm)} \\
\midrule
1 & 1 & Primary & $\leq 500$ & 35 & 2 & Primary & $\leq 2500$ \\
2 & 1 & Secondary & $\leq 500$ & 36 & 2 & Secondary & $\leq 2500$ \\
3 & 2 & Primary & $\leq 500$ & 37 & 3 & Primary & $\leq 2500$ \\
4 & 2 & Secondary & $\leq 500$ & 38 & 3 & Secondary & $\leq 2500$ \\
5 & 3 & Primary & $\leq 250$ & 39 & 4 & Primary & $\leq 1000$ \\
6 & 3 & Secondary & $\leq 250$ & 40 & 4 & Secondary & $\leq 1000$ \\
7 & 1 & Primary & $\leq 1000$ & 41 & 5 & Primary & $\leq 1000$ \\
8 & 1 & Secondary & $\leq 1000$ & 42 & 5 & Secondary & $\leq 1000$ \\
9 & 2 & Primary & $\leq 1000$ & 43 & 6 & Primary & $\leq 1000$ \\
10 & 2 & Secondary & $\leq 1000$ & 44 & 6 & Secondary & $\leq 1000$ \\
11 & 4 & Primary & $\leq 250$ & 45 & 1 & Primary & Any \\
12 & 4 & Secondary & $\leq 250$ & 46 & 1 & Secondary & Any \\
13 & 3 & Primary & $\leq 500$ & 47 & 2 & Primary & Any \\
14 & 3 & Secondary & $\leq 500$ & 48 & 2 & Secondary & Any \\
15 & 5 & Primary & $\leq 250$ & 49 & 3 & Primary & Any \\
16 & 5 & Secondary & $\leq 250$ & 50 & 3 & Secondary & Any \\
17 & 3 & Primary & $\leq 1000$ & 51 & 4 & Primary & $\leq 1500$ \\
18 & 3 & Secondary & $\leq 1000$ & 52 & 4 & Secondary & $\leq 1500$ \\
19 & 6 & Primary & $\leq 250$ & 53 & 5 & Primary & $\leq 1500$ \\
20 & 6 & Secondary & $\leq 250$ & 54 & 5 & Secondary & $\leq 1500$ \\
21 & 1 & Primary & $\leq 1500$ & 55 & 6 & Primary & $\leq 1500$ \\
22 & 1 & Secondary & $\leq 1500$ & 56 & 6 & Secondary & $\leq 1500$ \\
23 & 2 & Primary & $\leq 1500$ & 57 & 4 & Primary & $\leq 2500$ \\
24 & 2 & Secondary & $\leq 1500$ & 58 & 4 & Secondary & $\leq 2500$ \\
25 & 3 & Primary & $\leq 1500$ & 59 & 5 & Primary & $\leq 2500$ \\
26 & 3 & Secondary & $\leq 1500$ & 60 & 5 & Secondary & $\leq 2500$ \\
27 & 4 & Primary & $\leq 500$ & 61 & 6 & Primary & $\leq 2500$ \\
28 & 4 & Secondary & $\leq 500$ & 62 & 6 & Secondary & $\leq 2500$ \\
29 & 5 & Primary & $\leq 500$ & 63 & 4 & Primary & Any \\
30 & 5 & Secondary & $\leq 500$ & 64 & 4 & Secondary & Any \\
31 & 6 & Primary & $\leq 500$ & 65 & 5 & Primary & Any \\
32 & 6 & Secondary & $\leq 500$ & 66 & 5 & Secondary & Any \\
33 & 1 & Primary & $\leq 2500$ & 67 & 6 & Primary & Any \\
34 & 1 & Secondary & $\leq 2500$ & 68 & 6 & Secondary & Any \\
\bottomrule
\end{tabular}
\end{table}

\end{document}